# Physics-informed mixture of experts network for interpretable battery degradation trajectory computation amid second-life complexities

Xinghao Huang [1], Shengyu Tao [1, 2, ]*, Chen Liang [1], Jiawei Chen [3], Junzhe Shi [2], Yuqi Li [4],
Bizhong Xia [1, ]*, Guangmin Zhou [1, ]*, Xuan Zhang [1, ]*

[1] Tsinghua Shenzhen International Graduate School, Tsinghua University, Shenzhen, 518055, China
[2] Department of Civil and Environmental Engineering, UC Berkeley, Berkeley, CA, 94720, USA
[3] School of Computer Science, Peking University, Beijing 100871, China
[4] Department of Materials Science and Engineering, Stanford University, Stanford, CA, 94305, USA

* Corresponding authors:
sytao@berkeley.edu (S. Tao); xiabz@sz.tsinghua.edu.cn (B. Xia),
guangminzhou@sz.tsinghua.edu.cn (G. Zhou); xuanzhang@sz.tsinghua.edu.cn (X. Zhang).

## Highlights:

1. Physics-informed mixture of experts network for battery degradation trajectories computation.
2. Adaptive multi-degradation prediction module for latent degradation trend representation.
3. Future use-dependant degradation trajectories computation within a millisecond per battery.
4. Interpretability, generalizability and accessibility for safe and sustainable second-life uses.

## Broader context

As electric vehicle batteries reach end-of-life, their residual capacity presents substantial opportunities for second-life applications. However, ensuring safe and sustainable reuse remains challenging due to the absence of historical data and the uncertainty of future operating conditions. Conventional models often rely on consistent use profiles and complete cycling histories, which are rarely available in real-world practice. This study introduces a Physics-Informed Mixture of Experts (PIMOE) framework that classifies internal degradation modes from partial, field-accessible signals, such as voltage relaxation and partial charging data, captured at random states of charge regions. Integrating assumed future use scenarios, PIMOE enables accurate, long-horizon degradation trajectory computation without prior cycling data records. Its interpretable architecture reveals latent degradation modes that correspond to external trajectory geometry, serving as initial points for iterative degradation trajectory computation in extended second-life uses. This physics-informed approach bridges data-driven modeling with domain knowledge, offering a scalable, explainable, and computationally efficient tool for second-life battery deployment. The PIMOE holds broad implications for proactive fault detection, reliability assessment, and the durable operation of critical infrastructures such as grid-scale energy storage systems and massive electric vehicles.

## Keywords

# Abstract

Retired electric vehicle batteries offer immense potential to support low-carbon energy systems, but uncertainties in their degradation behavior and data inaccessibilities under second-life use pose major barriers to safe and scalable deployment. This work proposes a Physics-Informed Mixture of Experts (PIMOE) network that computes battery degradation trajectories using partial, field-accessible signals in a single cycle. PIMOE leverages an adaptive multi-degradation prediction module to classify degradation modes using expert weight synthesis underpinned by capacity-voltage and relaxation data, producing latent degradation trend embeddings. These are input to a use-dependent recurrent network for long-term trajectory prediction. Validated on 207 batteries across 77 use conditions and 67,902 cycles, PIMOE achieves an average mean absolute percentage (MAPE) errors of 0.88% with a 0.43 ms inference time. Compared to the state-of-the-art Informer and PatchTST, it reduces computational time and MAPE by 50%, respectively. Compatible with random state of charge region sampling, PIMOE supports 150-cycle forecasts with 1.50% average and 6.26% maximum MAPE, and operates effectively even with pruned 5MB training data. Broadly, PIMOE framework offers a deployable, history-free solution for battery degradation trajectory computation, redefining how second-life energy storage systems are assessed, optimized, and integrated into the sustainable energy landscape.

# Introduction

Lithium-ion batteries (LIBs), the core energy storage medium for electric vehicles and renewable energy systems, are experiencing surging growth in production and retirement [1–3]. It is projected that 120 GWh of battery capacity will reach end of life (EOL) by 2030, and the reuse and recycling of retired batteries has been identified a critical pathway for affordable, accessible and sustainable battery lifecycles, particularly for underdeveloped regions [4–7].

Capacity fade is a widely adopted performance indicator of battery degradation that can be relevant to the safety in extended use [8–11]. Retired batteries undergo capacity and consistency screening before second-life deployment to meet less-demanding application requirements [12,13]. However, uncertain use conditions in second-life scenarios significantly differ from first-life service patterns, rendering conventional capacity computational models ineffective, given that the data used for model training were built on constant-stress tests. The unavailability of historical cycling data for retired batteries complicates the tracing of coupled degradation modes, such as impedance rise (IR), loss of lithium inventory (LLI), and loss of active material (LAM) [14–17]. These degradation modes exhibit nonlinear interactions depending on prior use patterns [18–21], complicating the future degradation trajectory computation. Traditional non-computational methods relying on destructive disassembly or full-capacity calibration tests [22,23], inflicting extra damage and consumes extra larbor investments. For safety and sustainability awared reuse allocation at scale, such methods are impractical due to prohibitive time, momentary, and environmental costs [1,24–26].

Recent efforts focused on top-down physics-based models [9,27] (e.g., equivalent circuit models or electrochemical state-space models) to characterize degradation by simulating voltage, current, and temperature behaviors. Yet, these approaches struggle with parameter variations and calibrations for retired batteries due to the limited adaptability to uncertain use conditions in extended second-life. While promising for establishing statistical correlations between historical features [28–32] (e.g., voltage, resistance, and capacity based) and degradation, Tang *et al.* employed historical capacity data to compute degradation trajectories with domain shift conditions via meta-learning [33]; however, this approach harbors implicit assumption flaws, as it relies on curve fitting under constant use conditions by presuming sufficient historical data and known future use pattern. Tao *et al.* leveraged the physics-informed machine learning to reduce the data requirements to achieve entire degradation trajectory prediction using early data, however, it still requires historical data despite the amount is small [34]. In contrast, *Lu et al.* leveraged at least one full cycle of capacity-voltage curves to compute degradation trajectories under uncertain future operating scenarios by learning the relationship between use condition and battery capacity [35]. Yet, their method encounters limitations due to the stochasticity in state of charge (SOC) of retired batteries and the demanding requirement for full discharge-charge data. Notably, the demanding monitoring and recording processes are necessary for such data during the first-life service phase, but second-life battery stakeholders cannot afford such information given considerable customer privacy concerns and data inaccessibility, despite European battery passports initiatives and other data recording services [36–38]. Tao *et al.* proposed an state of health (SOH) estimation method for second-life batteries compatible with randomized SOC conditions, but it computes one capacity point instead of future battery degradation trajectory [12]. Thus, robustly computing future degradation trajectories using field-available signals, with randomly initialized SOC condition and under uncertain second-life use complexities remains a critical challenge.

This work proposes a Physics-Informed Mixture of Experts (PIMOE) network for computing degradation trajectories of retired batteries under uncertain second-life use complexities without requiring historical data. As shown in **Fig. 1a**, the model uses electrical signal data from randomly sampled SOC regions at the collection moment, interatively incorporating assumed future use conditions to forecast degradation trajectories toward extend future horizons. **Fig. 1b** illustrates the designed Adaptive Multi-mode Degradation Prediction (AMDP) module, which integrates multiple expert systems' weight that studied from capacity-voltage and relaxation data to achieve classification of degradation modes. AMDP outputs latent degradation trend embedding vectors that are forwarded into an Feature-Operational recurrent neural network (FORNN), enabling interpretable and extendable computation of degradation trajectories. In **Fig. 1c**, the model's application implicaiton on battery reuse and recycling are illustrated, highlighting an informed allocation of second-life batteries based on their computed degradation trajectories for a sustainable battery circular economy. Without historical data, our approach achieves an average MAPE of 0.88% in computing degradation trajectories under uncertain and diversified second-life use scenarios, spanning 207 batteries and 73 second-life use conditions. The required input data can be easily extracted at random SOC region, without charging or discharging batteries to anticipated SOC levels. The method can be extends to long-term degradation trajectory computation over a 150 cycle horizon with a MAPE of 1.50%. This study underscores transformative potential of embedding physics-informed principles into computational intelligence, enabling proactive fault detection and reliability assessment across complex, safety-critical systems, such as energy storage infrastructures, with implications for their durability, safety, and sustainable operation.

## Methodology overview

The objective of this study is to compute battery degradation trajectories under uncertain future use conditions using one cycle data collected at random SOC regions. We use partial charging curves, relaxation voltage, and future conditions (charge/discharge current, temperature) as required model inputs (specific extraction methods are detailed in **Method**). Input signal selection prioritizes two critical aspects: first, the controllability of charging processes in second-life battery ensures easier signal acquisition; second, relaxation voltage measurements remain unaffected by charge/discharge operations, thus enhancing model applicability. Six physics-informed features extracted from both relaxation voltage and partial charging curves (detailed in **Supplementary Figs. 1-3 and Supplementary Notes 1-2**) characterize polarization and degradation states. These features drive the model to classify batteries with different aging modes into distinguishable latent subspaces. Notably, physics-informed feature extraction extends beyond the charging and relaxation voltage curves discussed here, but serves as the inputs of the AMDP module.

A two-phase collaborative mechanism achieves deep integration of physical insights and data-driven approaches: (1) AMDP, and (2) FORNN. The AMDP module employs "degradation-router" routing to assign expert network weights based on the physical features, adaptively modeling degradation modes under dominant mechanisms for degradation trend representation[39], which can be formulated as:

$$Trend_i = \sum_{j=1}^{N_{expert}} G_j(\boldsymbol{F}_i) \cdot \text{Expert}_j(\boldsymbol{Q}_i) \qquad (1)$$

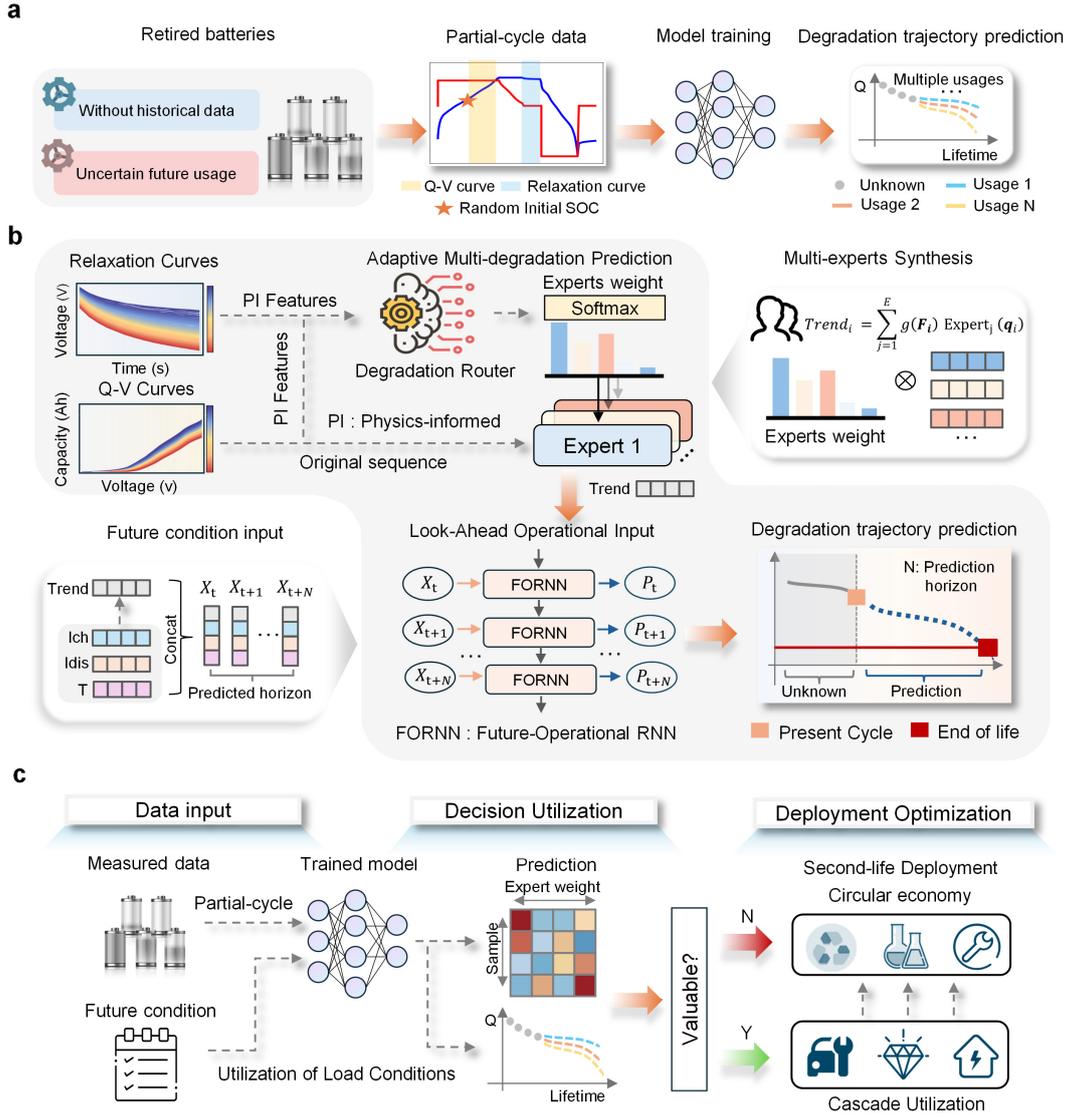

**Fig. 1 Model motivation, model architecture, and model deployment. (a)** Using field-accessible partial cycle data with random initial SOC to predict retired battery degradation trajectories under conditions of unknown historical data and uncertain secondary use condition. **(b)** The PIMOE model consists of two modules: AMDP and FORNN. The AMDP module uses physical features extracted from relaxation voltage and random initial SOC charging curves to guide the degradation router in adaptively integrating predictions from multiple degradation modes to output degradation trends. The FORNN module combines the degradation trends predicted by the AMDP module with future-operational to predict future degradation trajectories. **(c)** The trained model can utilize the output expert network to perform value assessment and preliminary utilization decisions for retired batteries and conduct economically and temporally reasonable secondary utilization deployment and recycling.

where, $Trend_i$ denotes computed degradation trend vector for the $i$-th battery sample, $F_i$ represents the physics-informed features extracted from the field cycle, and $Q_i$ is the partial charging curve (possibly with random initial SOC) for the field cycle. $N_{expert}$ is the total number of expert networks, and $G_j(F_i)$ is the router weight associated with $Expert_j$, determined by degradation-router mechanism. $Expert_j(Q_i)$ denotes the specific degradation trend estimator under the $j$-th dominant degradation mode.

The FORNN module iteratively integrates AMDP predictions regarding latent degradation trend with assumed future load conditions cycle-by-cycle, enabling proactive degradation trajectory computation without historical data, which can be formulated as:

$$\widehat{S}_i = \text{FORNN}(Trend_i, \boldsymbol{Cond}_i) \qquad (2)$$

where, $\widehat{S}_i$ represents the final predicted degradation trajectory (e.g., capacity fade curve) for the i-th battery sample over the computation time horizon, $Trend_i$ is the short-term trend output from AMDP module, and $\boldsymbol{Cond}_i$ contains future usage conditions such as charge/discharge current, which can be assumed as available. The FORNN leverages the sequential modeling capability to capture how future load profiles affect the long-term degradation starting from the current state.

These two modules exhibit unique but complementary strengths that AMDP computes latent degradation trends from learned degradation modes while FORNN deals with second-life use complexities (see **Supplementary Fig. 7** for the detailed machine learning pipleine). The degradation routing simultaneously considers physics-informed features reflecting current existing degradation modes and future load profiles to select optimal experts, forming a closed-loop physics-driven specialization and data-driven fusion architecture. Theoretically, PIMOE's design philosophy benefits from the success of collaborative learning [1,40], where weighted collaboration from multiple networks enhances prediction robustness [41–43]. In principle, PIMOE applies to degradation trajectory computation across entire lifecycle stage under arbitrary use conditions, including performance computation in other lifetime stages desipte considerably different internal degradation modes.

## Results

### Dataset

This work addresses the computation challenge of degradation trajectories for retired batteries under uncertain future use conditions using only partial cycling data at test field, i.e., no historical data is assumed. To validate computation performance when historical and future operating conditions are consistent but historical data is unavailable, we selected the Uniform-Life (UL) dataset for numerical experiment. To validate computation performance when the second-life use conditions change significantly as compared to their first-life counterpart, we selected the Two-Phase Second-Life (TPSL) dataset for numerical experiment.

The UL dataset comprises batteries tested under different use conditions, with each batch cycled under consistent conditions throughout their entire lifespan until EOL, comprising three batches totaling 130 commercial 18650 cells. Batch 1 consists of NCA batteries with 3500mAh nominal capacity and cutoff voltages of 2.65–4.2 V. Batch 2 contains NCM batteries with 3500mAh nominal capacity and cutoff voltages of 2.5–4.2 V. Batch 3 includes NCM+NCA hybrid batteries with 2500mAh nominal capacity and cutoff voltages of 2.5–4.2 V. All cells underwent cycling in thermal chambers under three temperatures (25°C, 35°C, 45°C) with variable charge rates (from 0.25C to 4C) and fixed 1C discharge rate. Each battery experienced identical operational profiles throughout its full lifecycle until reaching EOL (see **Supplementary Table 1** for details). Degradation trajectories in **Fig. 2a** demonstrate cycling variability, while **Fig. 2b** illustrates significant remaining useful life (RUL) differences among batteries.

The TPSL dataset contains batteries subjected to diverse second-life scenarios, where they undergo both randomized and standardized operating conditions following their primary usage phase, containing 77 pecices of 18650 batteries with LiCoO$_2$ and LiNi$_{0.5}$Co$_{0.2}$Mn$_{0.3}$O$_2$ cathodes and graphite anodes. These 2.4 Ah batteries (3.7 V nominal, from 3.0 to 4.2 V cutoffs) underwent two-phase testing: Phase 1 involved 20 cycles of 0.5C constant-current constant-voltage (CCCV) charging and 2C discharging. Phase 2 divided batteries into two subgroups: (1) TPSL-Random (55 cells) subjected to stochastic charging profiles with current rates (1C/2C/3C) changed randomly every 5 cycles that follow a uniform statisitical distribution, combined with fixed 3C discharge; (2) TPSL-Fixed (22 cells) cycled under predetermined charge/discharge current combinations (1C/2C/3C) (details in **Supplementary Table 2**). **Fig. 2c** shows degradation trajectories under uncertain use conditions, disproving assumption of identical operational history for prediction reliability. **Fig. 2d** reveals substantial capacity dispersion at cycles 21, 60, and 100 post-load variation, demonstrating how uncertain loads interacting with battery degradations. **Fig. 2e** details the input data construction process, where each sample combines raw charging curve sequences with physics-informed features, see **Method** (data processing) and **Supplementary Note 2**.

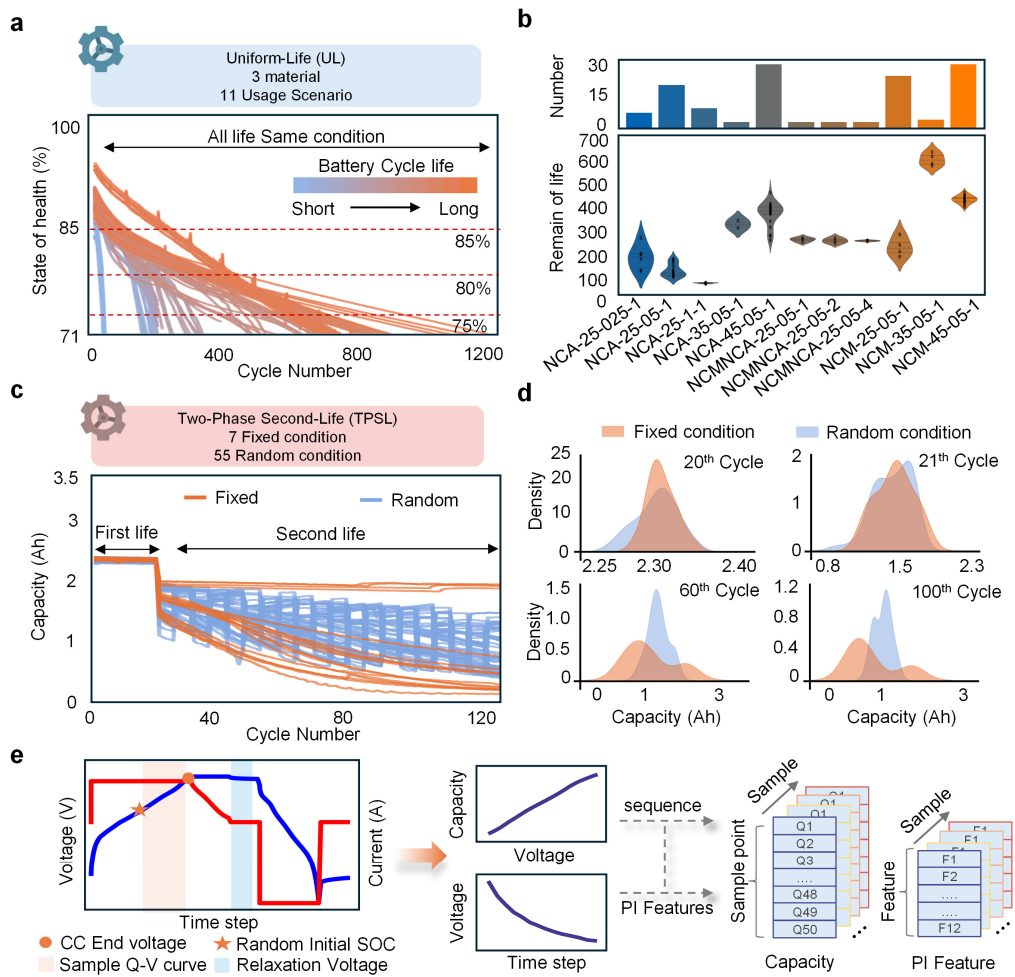

**Fig. 2 Dataset Description.** (a) Summary of the UL dataset testing conditions and the degradation curves of the batteries. (b) Remains useful life distributions of the cells under identical operating conditions remain (naming convention refers to **Supplementary Tables 1-2**). (c) Summary of the TPSL dataset testing conditions and the degradation curves of the batteries. **(d)** Capacities distributions of the cells at different use condition. **(e)** The charging curves with random initial state of charge and the 30-minute relaxation voltage curves.

**Model performance and generalization capability**

To validate the robustness of the proposed method under unknown historical data and uncertain second-life use conditions, we conducted evaluations in three real-world application scenarios (UL, TPSL-Random, TPSL-Fixed). PIMOE adaptively computes nonlinear degradation mode under uncertain use condition through AMDP module, see **Supplementary Table 3** for the accociated model parameter.

Performance metrics (RMSE, MAPE, $R^2$) are in **Method** (Evaluation metric). It is emphazied that all experiments used current-cycle data with random initial SOC regions extracted from field conditions to compute degradation trajectories spanning dozens to hundreds of cycles, rather than EOL points (see the prediction horizon selection in **Supplementary Note 3**). As a time-series computation task, we benchmarked the model performance against state-of-the-art (SOTA) sequence computation models [44,45] PatchTST and Informer (see detail in **Supplementary Note 4** and **Supplementary Table 3** for detailed information). In **Fig. 3a**, existing research demonstrates that battery degradation can be divided into three typical while distinct degradation phases: SEI formation, SEI thickening, and lithium plating [8,46]. The degradation patterns across these phases exhibit significant differences in both available capacity and subsequent aging rates, geometrically the slope of degradation trajectory [38,47]. We randomly selected representative samples from each degradation phase to analyze performance, where each sample reflects distinct dominant degradation modes with varying capacity retention and aging rates. While PatchTST and Informer showed errors across early, middle, and late degradation phases, PIMOE demonstrated superior stability, delivering accurate current-cycle capacity estimation while generating trajectory predictions consistent with phase-specific degradation rates. **Supplementary Figs.8-10** validate PIMOE's effective prediction capability across full lifecycle samples.

**Fig. 3b** reveals that in TPSL data, baseline methods displayed limited capability in capturing long-term trends after future load changes and completely failed during initial load transitions. In contrast, PIMOE effectively identified transitional characteristics between phases through future usage condition integration. **Supplementary Figs.11-14** further demonstrate PIMOE's successful learning of relationships between future operating conditions and capacity performance.

**Fig. 3c** demonstrates that PIMOE exhibits significant and stable advantages in both MAPE and $R^2$ metrics with the cross-dataset performance comparison setting. For UL dataset, PIMOE achieved its worst prediction result on the NCA-25-1-1 dataset with an average MAPE of 0.98%, while maintaining MAPE below 1.00% across all UL conditions, showcasing robust prediction performance. PatchTST delivered the second-best results with an average MAPE of 0.85% and a maximum MAPE of 1.55% in the UL dataset, indicating that its patch-based tokenization strategy captures aging variations more effectively than single-point sampling. Informer performed the worst in this scenario, with an average MAPE of 1.48% and a maximum MAPE of 2.13%, suggesting that its sparse attention mechanism may lose critical information and fail to detect subtle temporal differences in voltage-capacity curves during such extreme prediction tasks. In the TPSL dataset, PIMOE showed even greater performance advantages over baseline methods, achieving an average RMSE of 0.05 compared to 0.38 for PatchTST and 0.39 for Informer, further confirming the necessity of incorporating future operating conditions in second-life applications.

In **Fig. 3d**, the scenario analysis confirms that PIMOE achieved an average MAPE of 0.52% with a standard deviation of 0.28% for the UL dataset, 2.96% with a standard deviation of 0.44% for the TPSL-Fixed dataset, and 2.81% with a standard deviation of 0.15% for the TPSL-Random dataset. PIMOE performed better on UL data with continuous operation history than on TPSL data with varying conditions, primarily because continuous operation provides more stable degradation pattern characteristics, enabling the model to more accurately capture battery degradation patterns. In contrast, random load switching in TPSL data introduces additional cycle-level nonlinear complexities that increased degradation trajectory computation difficulty. Nevertheless, all MAPE values remained below 3.00%, demonstrating excellent performance in computing degradation trajectories under uncertain future conditions from current-cycle data without requiring historical records.

In **Fig. 3e**, we analyzed the relationship between model computational efficiency and performance to meet the real-world deployment requirements. After training, PIMOE required 0.43ms to infer the entire degradation trajectory in the coming hundreds of cycles incorporating future operating conditions for a single retired battery, while achieving an MAPE of 0.88%. In comparison, baseline methods PatchTST and Informer required 0.58ms and 0.87ms per battery, with MAPEs of 4.29% and 4.89%, respectively. PIMOE thus achieves the most robust prediction accuracy while emphasizing lightweight real-world deployment.

Unlike existing approaches that rely on fixed voltage or SOC regions and require time-consuimng SOC recalibration, we validated PIMOE's robustness across five random initial SOC regions. In **Fig. 3f**, model prediction performance was relatively poorer at an initial SOC of 50%, with MAPEs of 1.44%, 3.35%, and 3.43% for the UL, TPSL-Random, and TPSL-Fixed datasets respectively, accompanied by standard deviations of 0.84%, 0.14%, and 0.50%. Reducing the initial SOC to 20% improved average performance by 63.8%, 16.2%, and 13.7% respectively. This performance difference likely occurs because higher initial SOC reduces available training data volume, thereby decreasing model performance. Additionally, since UL dataset computations do not involve any cycle-level condition changes, they depend more heavily on the data volume available at the initial SOC, resulting in more pronounced performance degradation as initial SOC increases.

**Rationalization of statistical model performance**

Here we explore the fundamental mechanisms underlying performance improvements. In **Fig. 4a**, we first randomly selected NCA battery test samples from the UL dataset to visualize the weight assignments of the degradation router throughout their full lifecycle. Without requiring historical data, the degradation router partitions battery lifecycle degradations into three distinct phases using partial cycling data, which is evidenced by existing literature [8,46]. It is observed that early-retirement samples are predominantly governed by Expert Networks 1 and 4, mid-life retired batteries exhibit higher weights for Expert Networks 2 and 3, while late-life retired batteries are primarily regulated by Expert Network 5. These results demonstrate PIMOE's capability to identify evolving degradation patterns, with clear functional specialization among expert networks for different degradation stages. Mechanistically, this suggests

Expert Networks 1, 3, and 5 respectively dominate initial SEI layer formation, subsequent thickening, and lithium plating processes, consistent with prior research on primary battery degradation modes [46]. Expert Network 4 maintains high weights during both early and late degradation phases, potentially reflecting its statistical correlation with high degradation rates, i.e., slope of the degradation trajectories, rather than specific mechanistic contributions.

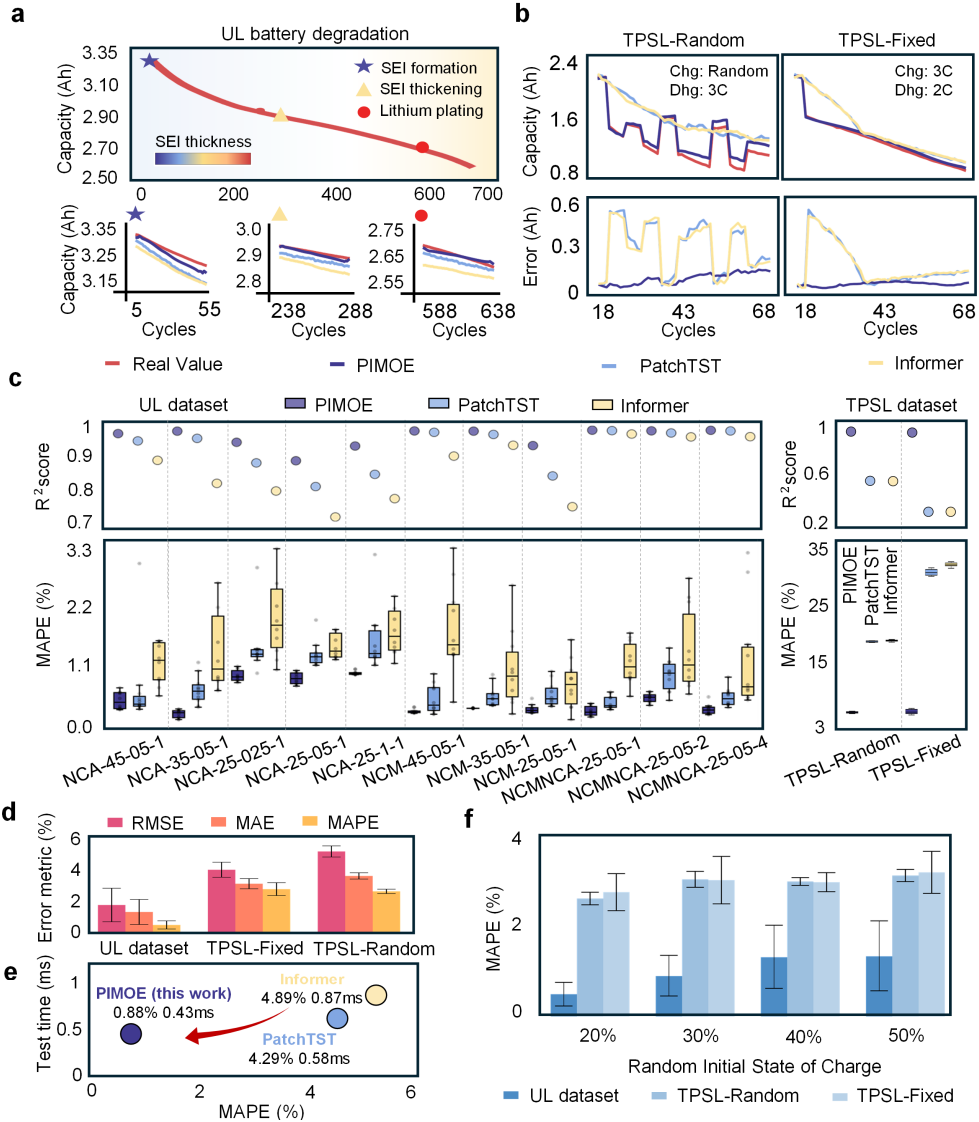

**Fig. 3 PIMOE results and analysis. (a)** Prediction results of the proposed framework and existing methods for interval samples dominated by three different degradation modes. **(b)** Prediction results of the proposed framework and existing methods under different secondary use condition. **(c)** Comparison of prediction results for different materials and usage conditions. **(d)** Prediction results for datasets under three different usage conditions. **(e)** Comparison of prediction time and accuracy between the proposed framework and existing methods. **(f)** Model performance under different initial SOC conditions in random retirement scenarios.

**Fig.4b** takes Battery 1 as an example to deeply analyze the dynamic correlation between individual cycling samples and expert network weight assignments during capacity degradation. Single-cycle samples at similar degradation stages exhibit clustering characteristics in expert weight allocation, demonstrating highly consistent expert category selection. Meanwhile, the degradation router adaptively adjusts the weights of different expert networks based on the features of individual cycling samples, thereby achieving differentiated prediction of degradation variations both between cycles and across degradation stages. **Supplementary Fig. 15** further confirms the observed expert weight dependencies across different battery material systems and second-life use conditions.

We performed t-SNE dimensionality reduction experiments on expert weights across the entire test battery cohort (details in **Supplementary Note 8**) to visualize the separability of decision boundaries among expert networks. In **Fig. 4c**, despite significant inter-cell variability, samples from early and late degradation stages form distinct decision boundaries, as demonstrated in **Supplementary Figs. 19-21** that PIMOE has generalizability across battery material systems and operating conditions. These findings raise a critical question: can the trained PIMOE model directly classify retired batteries based on degradation router-generated weight patterns? In essence, this question is a interpretability analysis showing why AMDP should work well.

In **Fig. 4d**, for 130 UL dataset batteries retired at different SOH levels (classification criteria in **Supplementary Note 7**), those retired at 95% SOH achieved "excellent" classification confidence (91.5%), while 75% SOH batteries received "scrap" ratings (96%). **Supplementary Figs. 16-18 and Supplementary Table 11** visualize the expert weight distributions across different SOH, with marked differences further demonstrating our proposed method's novelty and simplicity of AMDP module that can classify the degradation mode using a few electric measurements. **Fig. 4e** examines post-classification computation accuracies of degradation trajectories . PIMOE achieves MAPEs of 0.47%, 0.52%, and 0.47% for batteries retired at 95%, 85%, and 75% SOH, respectively. While predictions remain robust for 95%-85% SOH, samples at 75% SOH show maximum outliers, likely due to coupling of accumulated and other unobserved degradation mechanisms.

In **Fig. 4f,** controlled experiments with noise addition validate feature importance and expert network specialization. By individually applying Gaussian noise ($\sigma = 1$) to each physical feature while holding other feautures at their original value, we observe elevated errors from baseline MAPE of 1.13% when features are perturbed with noise. With the same amount of noise injection, the perturbed feature leads to the most decreased degradation trajectory computation accuracy is regarded as the most important. It reveals that features extracted from relaxation voltage that is reflective of internal resistance, demonstrate greater noise resilience than those from capacity-voltage curves (particularly their mean values, after adding noise, the MAPE increased from 1.13% to 1.38%, representing a 22% relative increase. as these means directly reflect charge storage capacity per voltage increment). The observed performance decrease caused by misrouting samples to expert networks not specialized for their specific degradation phases (e.g., late-stage samples erroneously routed to Expert Network 1 instead of 5) confirms the functional divergence among expert networks, enabling history-free computation of degradation mode. **Supplementary Figs. 22-24** show reduced specialization when degradation trajectories are geometrically similar，while demonstrating that PIMOE's superior performance does not strictly rely on any single physical feature.

In **Fig. 4g**, we investigate the contribution of the AMDP module to prediction accuracy using an ablation experiment. When replacing AMDP with a simple linear layer while keeping other components unchanged, comparative analysis of the embedded representations (see **Supplementary Note 6** for details) shows that while the ablated model can roughly distinguish degradation stages, it fails to capture subtle SOH variations within similar stages. This results in discontinuous embeddings and unreasonable trajectory predictions, highlighting AMDP's critical role in enabling robust long-term predictions under history-free settings. By decoding implicit semantics of expert weights, PIMOE transforms from a prognostic tool into an interpretable decision engine, bypassing traditional sequential assessment workflows that requires extensive data curation and model calibration in real-world deployment.

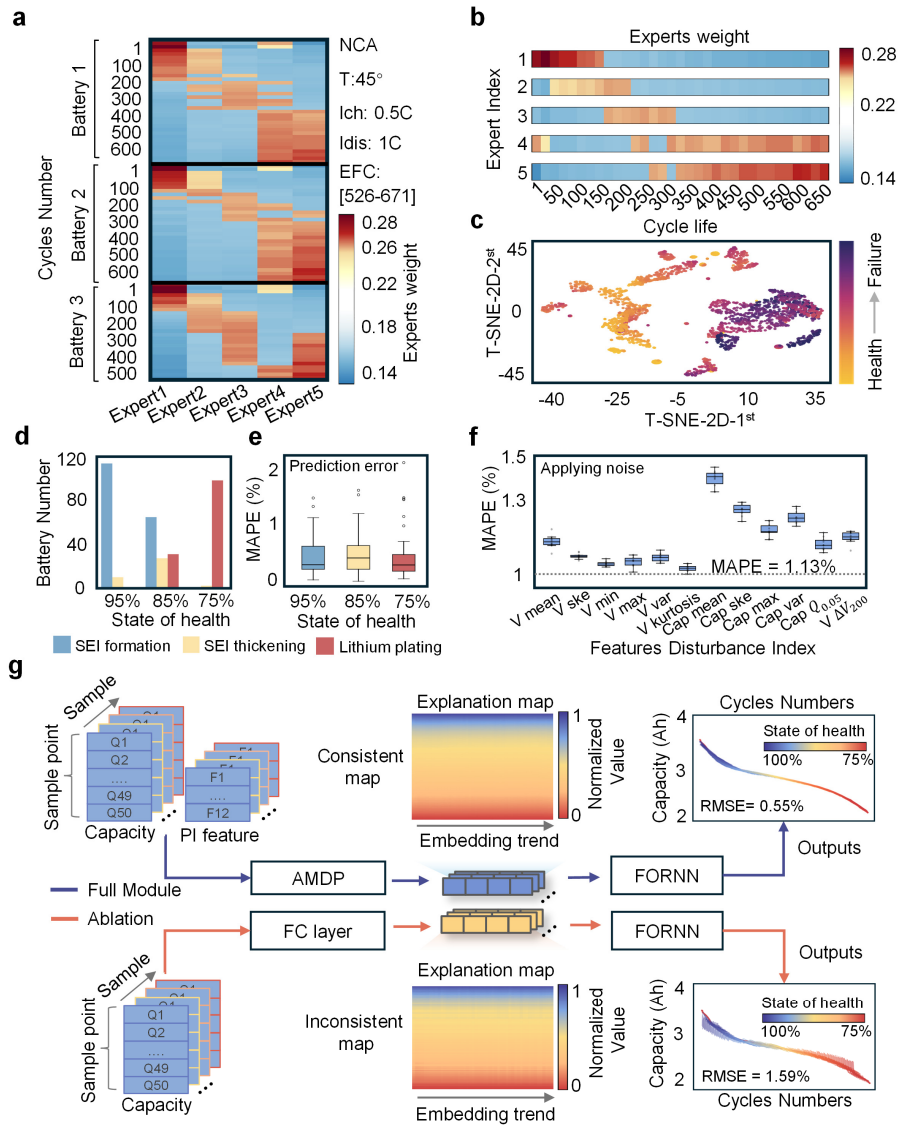

**Fig. 4 PIMOE interpretability. (a)** Visualization of degradation expert weights for selected NCA batteries. **(b)** Degradation expert weights corresponding to different intervals in random retirement scenarios throughout battery lifecycle. **(c)** Visualization of expert weights after T-SNE dimensionality reduction to 2D. **(d)** Classification results of UL dataset batteries under different SOH retirement scenarios using expert weights only (specific classification scheme see **Supplementary Note 7**). **(e)** Prediction results for batteries retired at different SOH levels. **(f)** Prediction results under noise injection with varying physical features. **(g)** Internal architecture and signal analysis diagram of the PIMOE.

**Uncertainty Analysis**

Under the assumption of available historical data and consistent second-life operational conditions, we extend our analysis to scenarios where partial historical data is accessible. While this significantly simplifies the prediction task, it does not align with the complex data composition of real-world applications, Using polynomial fitting and MLP-based approaches to map historical trajectories to future degradation (see the implementation details in **Supplementary Note 9**). **In Fig. 5a**, without requiring battery-specific mechanistic analysis, a simple polynomial regression and single-layer MLP achieved a 0.79% and 0.93% MAPE, respectively. We demonstrate the PIMOE's capability in computing degradation trajectories using historical data regarding maximum capacity to validate its adaptability to data- accessible scenarios (see **Supplementary Fig. 25 and Supplementary Note 11** for implementation details), With 10 historical maximum capacity points accessible, PIMOE achieves an average MAPE of 0.53% when predicting the next 50 cycles. We stress that these historically-dependent methods become infeasible when historical data is unavailable in real-world settings, underscoring the practical limitations of such assumptions. To quantify component contributions of AMDP and FORNN, we designed two ablation studies: (1) replacing AMDP with a linear layer (PIMOE-linear); and (2) substituting FORNN with a standard RNN (PIMOE-woFO). After replacing the AMDP architecture with a linear layer directly, the overall MAPE increased from 0.88% to 1.14%, which aligns with the observation in **Fig. 3e** that the embedding vectors became indistinguishable between different degraded samples after removing AMDP architecture. This demonstrates that the adaptive integration of predictions from different degradation modes contributes to robust and high-precision estimation. When replacing FORNN with a standard RNN, the model exhibited a significant decline in prediction performance for the TPSL dataset. This further underscores the necessity of incorporating future load conditions in second-life scenarios, as future operating conditions often differ from those in their first-life service patterns.

**Fig. 5b** evaluates the PIMOE performance under limited training data access (data partitioning details are provided in **Supplementary Note 7** and **Supplementary Table 10**). When training data was reduced from 100% to 40%, the model performance maintained stable on UL dataset, achieving an average MAPE of 0.75% with a 0.56% standard deviation. However, a linear performance decrease was observed on TPSL dataset, with average MAPE increasing to 3.86% with a 0.88% standard deviation. When training data was reduced to 20% (32 batteries in total, pruned training data size 5MB), both datasets exhibited noticeable error and uncertainty increase, with MAPE reaching 1.07% for UL and 5.30% for TPSL. This training data access experiment indicates that learning future operating conditions' impact on battery degradation trajectories requires higher data curation costs, whereas the UL dataset demonstrates relatively lower data access dependency. Desipte this observed data access challenge, **Supplementary Tables 5-7** compare baseline model performance under scarce data access, validating PIMOE's superior generalization capability in few-shot learning scenarios.

The decomposition of degradation subspaces and dominant mechanism prediction depend on the number of experts and TopK selection (see **Methods**). **Fig. 5c** examines the impact of number of experts on the PIMOE performance. The PIMOE demonstrates relative insensitivity to hyperparameter modifications, across experiments with five distinct expert network quantities, the average MAPE was 0.98%. when employing 4 experts, the model showed comparatively poorer performance with an average MAPE of 1.03%, while the 6-expert configuration exhibited inferior stability with a standard deviation of 1.04%. After carefully balancing practical deployment requirements against performance metrics, we ultimately adopted a configuration with 5 experts and TopK=2.

Different application scenarios have varying requirements for prediction horizons. In practical applications, it is desirable to predict degradation trajectories as far as possible while maintaining acceptable performance. As shown in **Fig. 5d**, we present results across different prediction lengths. When the prediction length is reduced, model performance improves, achieving an average MAPE of 0.80%. In most cases, model performance degrades with increasing prediction length. Nevertheless, even when predicting 150 future cycles from current-cycle data, the average MAPE remains below 1.50%, demonstrating PIMOE's robustness in long-term forecasting.**Supplementary Tables 9-10** compare our method's performance with baseline approaches under extended prediction horizons. The results show that all models exhibit performance degradation as prediction length increases. When predicting future 100-cycle degradation trajectories, PatchTST and Informer achieve average MAPEs of 8% and 7.98%, respectively, while PIMOE maintains a lower average MAPE of 1.47%, enabling more accurate long-term predictions.

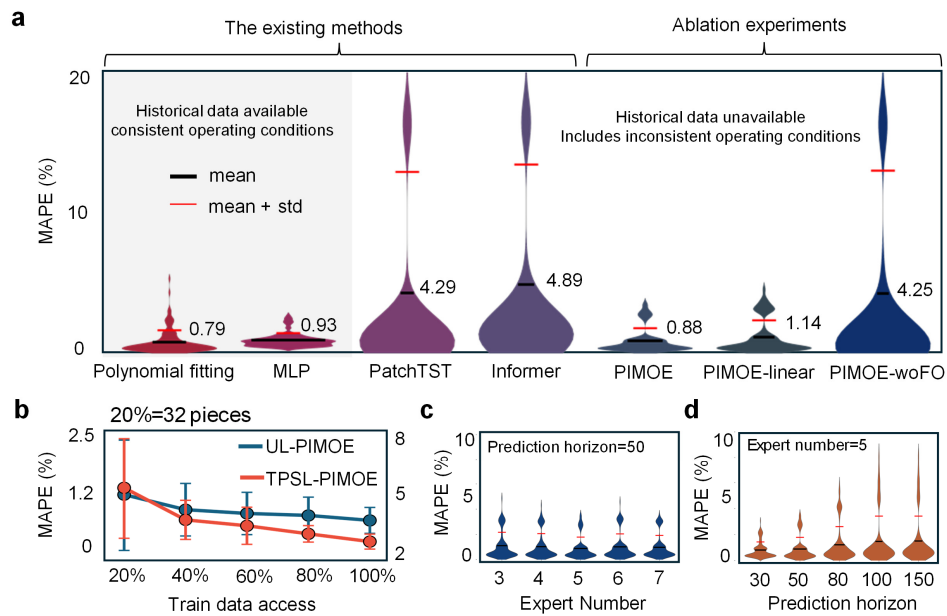

**Fig. 5 Uncertainty analysis. (a)** Comparison of MAPE error distributions for degradation trajectory predictions (gray areas represent conditions with available historical data and consistent secondary use conditions; blank areas represent comparisons with other models and ablation study results). **(b)** Model accuracy under limited training data conditions. Impact of **(c)** expert quantity and **(d)** prediction horizon on model accuracy.

## Discussion

Existing research paradigm of degradation trajectory computation predominantly relies on historical data access and assumed identical future use conditions, which is often not true for second-life batteries. This paradigm inflicts either expensive lifecycle data management or time-consuming capacity calibration tests post-retirement, raising safety concerns due to its inability to account for second-life use uncertainty. The proposed PIMOE overcomes these data accessibility limitations by enabling computation of degradation trajectories under uncertain future use conditions using data that is readily available in battery management systems, eliminating the need for additional offline testing. The success of PIMOE stems from adaptively and interpretably classifying degradation modes, statistically the geometric slope of the degradation trajectory at the observation point, to inform use-dependant degradation trajectory computation in the extended second-life use.

Through case studies involving three usage scenarios (i.e., UL, TPSL-Random, TPSL-Fixed), 203 independent cells, and 77 secondary-use operating conditions, we demonstrate that PIMOE achieves a promising lifecycle MAPE of 0.88% in computing degradation trajectories of 50 future cycles using only real-time online data, while reducing computation time by up to 50% compared to SOTA time-serials computational models. The average and maximum MAPE valuse across all datasets were reduced by 50% as compared to PatchTST and Informer. The PIMOE is flexible with fixed and random SOC regions, allowing for seamless integration into second-life batteries that exhitbit random initial SOC distributions. By adaptively integrating degradation mode classification using proposed AMDP module and assumed future use conditions, the battery degradation trajectory computation horizon can be extended to 150 cycles and beyond, with an average MAPE of 1.50% with a maximum MAPE below 6.26%. Hyperparameter sensitivity analysis confirms the model's robustness under uncertain second-life use conditions, with an average MAPE of 0.98%. The PIMOE interpretability is demonstrated via the visualization of latent embeddings from expert weights that represent degradation modes.

The PIMOE model shows great potential to transform the battery reuse and recycling industry landscape by reducing the need of manual-assisted testing approach to an automated data-driven decision support system. However, it must be acknowledged that the PIMOE approach primarily depends on electrochemical characteristics (such as voltage-capacity relationships and relaxation voltage) for macroscopic diagnostics. While the extensive validation confirms the physical understanding of features, they essentially remain statistical correlations, lacking systematic integration of deeper physicochemical mechanisms. It is recommented that future work should focus on the non-invasive *in-situ* sensing signals [2,48,49], such as vibration sensing, strain sensing, ultrasonic signals, fiber optic sensing to enhance internal state observation for a physics-driven modeling. Building a "physics-statistics" fusion framework has the potential to improve interpretability of PIMOE [30], for example, the parameters in the battery electrochemical or equivalent circuit models can be updated online as an intermediate physical interpretation layer for degradation trajectory computation [50]. Even though the assumption that future use conditions are available is reasonable for verifying the effectiveness of PIMOE, the future operating conditions of the battery energy storage system are difficult to predict in reality because they depend largely on typical application scenarios [19]. Although the paper has discussed the importance of introducing future use conditions in the FORNN structure, the analysis of the uncertainty of future use conditions needs to be extended to highly random and time-varying use scenarios, more importantly, with statistical confidence margins, of the battery energy storage system to ensure the safe and duration operation [7]. Although the paper has taken into account three very different application conditions, i.e., UL and TPSL-fixed and TPSL-fixed, considering that the second-life application of the battery is safety-critical, more extreme conditions need to be investigated, since the robustness of the dataset to non-steady-state use conditions, such as the battery approaching thermal runaway or internal short circuit is still not included. It is important to consider the bias of training data because UL and TPSL data hardly cover all environmental fluctuations under actual service conditions [39], thermal non-uniformity at battery pack level, internal resistance. Future work is suggested to collect real-world variables inclusive of different manufacturers [51], degradation mechanisms [16], and cell-to-cell variailites [52].

This work demonstrates the potential of history-free, proactive degradation trajectory computation amid second-life use uncertainty, reducing the need for time-consuming and costly offline testing while ensuring safety compliance. The PIMOE has demonstrated great poteintial in retired battery deployment decision-making at scale by laying a groundwork for light, generalizable battery management computational models. This work underscores the promise of integrating physics insights to computational models for early malfunction detection and reliability assessment, paving the way for safer and more durable operation of critical infrastructures such as energy storage systems.

# Method

## Data Preparation

**Input signals.** We consider two signals in each cycle: partial charging profile that starts at a random voltage (i.e., a random initial SOC) and ends at the cutoff voltage. Relaxation voltage curve recorded immediately after charging. Let $I(t)$, $V(t)$, and $t$ denote the instantaneous current, voltage, and time, respectively. We track the battery voltage from $V_{min}$, which varies across samples due to random SOC, up to the cutoff voltage $V_{end}$.

**Partial charging curve.** We divide voltage range $[V_{min}, V_{end}]$ into N increments of size $\Delta V$. The partial charging curve $q$ denotes the cumulative input charge as a function of voltage:

$$q = [q_0(V), q_1(V), \cdots, q_N(V)], \; q_i(V) = \int_{t(V_{min})}^{t(V_{min}+i\Delta V)} |I(t)| dt \tag{1}$$

In experiments, we record 50 segments at equal intervals (i.e., N = 50), yielding a 50-dimensional sequence.

**Relaxation voltage curve.** To capture the relaxation behaviour, we measure the open-circuit voltage $v_k(t)$ for $k = 0, 1, \ldots,$ N at 30-minute intervals after charging concludes:

$$v = [V_0(t) \; V_1(t) \; \cdots \; V_N(t)] \tag{2}$$

This sequence of length N + 1 reflects how the polarized voltage recovers over time.

**Feature engineering.** From partial charging and relaxation voltage signals, we extract 12 features derived from physical considerations, including polarization overcharge, relaxation slope, and voltage plateau transitions. All 12 features are normalized to [0,1] before entering the model. The final input to the PIMOE consists of: the partial charging curve $q \in \mathbb{R}^{N \times 50}$ and the feature vector $\boldsymbol{F_i} \in \mathbb{R}^{N \times 12}$ derived from both the charging and relaxation profiles.

## Model Architecture

The proposed Physics-Informed Mixture-of-Experts framework (PIMOE) consists of two modules, *Adaptive Multi-degradation Prediction* (AMDP) module and a *Future-Operation Recurrent Neural Network* (FORNN), to include current-cycle degradation phenomena and prospective operating conditions that may arise in second-life applications.

**Adaptive Multi-degradation Prediction (AMDP)**

Formally, let $\boldsymbol{F_i} \in \mathbb{R}^{N \times 12}$ be physics-informed features for the $i$-th sample, and $q = [q_0(V), q_1(V), \cdots, q_N(V)]$ be its partial charging curve. The AMDP model defines the expert networks, each denoted $\text{Expert}_j$, $j = 1, 2, \ldots, E$. A degradation mode routing $\boldsymbol{F_i} \mapsto g$ produces a probability weight $g(\boldsymbol{F_i})$ across these experts:

$$g(\boldsymbol{F_i}) = Softmax(Softmax(\text{TopK}(\text{H}(\boldsymbol{F_i}), k))) \tag{3}$$

where, $H(\mathbf{F}) \in \mathbb{R}^{N \times E}$ is a feed-forward transformation with a noise term:

$$\text{H}(\boldsymbol{F_i}) = \boldsymbol{F_i} \boldsymbol{W_g} + \psi \, \text{Softplus}(\boldsymbol{F_i} \boldsymbol{W}_{\text{noise}}) \tag{4}$$

where, $\boldsymbol{W_g} \in \mathbb{R}^{12 \times E}$ and $\boldsymbol{W_{noise}} \in \mathbb{R}^{12 \times E}$ are learned parameters, and $\psi \in N(0,1)$ is drawn from a standard

Gaussian. The top-$k$ experts are selected and normalized by a $Softmax$ function.

Retired batteries may undergo multiple degradation mechanisms, e.g., SEI formation, SEI thickening, loss of active material, and lithium plating, that can coexist or interact. To flexibly and specifically model these different pathways, AMDP module enables a Mixture-of-Experts (MOE) structure by fusing predictions from multiple expert submodels, each "specialized" in a certain degradation mode. Within AMDP enabled MOE framework, an Expert is not confined to a single degradation mode; instead, it is represented as a weighted superposition of countably many implicitly defined "typical degradation mode". Let the $j$-th $\text{Expert}_j$ be denoted as:

$$\text{Expert}_j(q_i) = \sum_{\theta \in \Omega} w_i(\theta)\, \phi(q_i; \theta) \tag{5}$$

where $q_i$ represents the partial charging sequence extracted under a random initial SOC condition. $\phi(q_i;\theta)$ is the basis response associated with the latent degradation mode indexed by $\theta$, $w_i(\theta)$ is the weight that $\text{Expert}_j$ assigns to corresponding degradation mode. The output is a short-term (within a small region around the computation point) degradation trend under the assumption that the $j$-th mechanism is dominant. The final AMDP output for sample $i$ is a weighted sum:

$$Trend_i = \sum_{j=1}^{E} g(F_i)\, \text{Expert}_j(q_i) : \mathbb{R}^N \to \mathbb{R}^L \tag{6}$$

$Trend_i \in \mathbb{R}^{1 \times L}$ represents the preliminary degradation trend output by AMDP module, where $L$ indicates length of the predicted future degradation trajectory. $\text{Expert}_j$ denotes the predictor for a specific degradation mode, and where $g_j(F_i)$ is the weight for expert $j$. This step produces a latent degradation trend vector $Trend_i$ for subsequent cycles, reflecting the dominant mechanism(s) underpinned by $F_i$. Hence, the model adaptively fuses multiple degradation mode subspaces into a short-term trend embedding for subsequent long-horizon computation.

**Future-Operation Recurrent Neural Network (FORNN)**

Second-life applications involve future load conditions that can alter degradation pace. To include these influences, we employ a recurrent neural network that takes as input AMDP output $Trend_i$ and a set of future load parameters:

$$C_i = \left[ \left(I_{\text{charge}}^{(1)}, I_{\text{discharge}}^{(1)}, T^{(1)}\right), \dots, \left(I_{\text{charge}}^{(L)}, I_{\text{discharge}}^{(L)}, T^{(L)}\right) \right] \tag{7}$$

where $C_i \in \mathbb{R}^{N \times L \times 3}$, $I_{\text{charge}}^{(\ell)}$, $I_{\text{discharge}}^{(\ell)}$, and $T^{(\ell)}$ represent the charge current, discharge current, and temperature in future cycle $\ell$. We concatenate these load vectors with the short-term trend from AMDP:

$$X_i = \text{Concat}(Trend_i, C_i) \tag{8}$$

The LSTM processes $X_i \in \mathbb{R}^{N \times L \times 4}$ sequentially to predict the capacity trajectory $\widehat{S}_i$ over $L$ future cycles:

$$\widehat{S}_i = \text{LSTM}(X_i) \tag{9}$$

By iterating cycle by cycle, FORNN accommodates a prediction horizon that the user chooses, merging the degradation modes signature from AMDP with the load sequence the battery will experience under the second-life complexities and uncertainties.

## Model Training

We train the AMDP and FORNN modules jointly by minimizing a loss function that balances trajectory accuracy with degradation router diversity. Let $S_i$ be the ground-truth capacity trajectory of length $L$ for sample $i$, and let $\widehat{S}_i$ be the corresponding model output. The trajectory fidelity objective is

$$\mathcal{L}_{\text{traj}} = \frac{1}{N} \sum_{i=1}^{N} \left\| \widehat{S}_i - S_i \right\|^2 \tag{10}$$

where the summation is over the training set of size $N$.

Because MOE can exhibit "expert collapse" (where one expert is used for all samples), we include a regularization term to encourage usage across experts. For a mini-batch, the router weight assigned to expert $j$ is averaged as

$$A_j = \sum_{i \in \text{Batch}} g_j(\boldsymbol{F}_i) \tag{11}$$

We encourage these averages to avoid large variation by penalizing their coefficient of variation:

$$\mathcal{L}_{router} = \frac{\text{Var}(A_1, A_2, \ldots, A_E)}{\left(\text{Mean}(A_1, A_2, \ldots, A_E)\right)^2 + \varepsilon} \tag{12}$$

where $\varepsilon$ is a constant **set to 10**, Var and Mean are the variance and mean operator, respectively. The total loss is:

$$\mathcal{L} = \alpha \, \mathcal{L}_{\text{traj}} + \beta \, \mathcal{L}_{router} \tag{13}$$

with $\alpha$ and $\beta$ controlling the trade-off between trajectory accuracy and expert diversity. We set $\alpha = 0.75$ and $\beta = 0.25$ in our experiments.

## Evaluation Metrics

### Root Mean Square Error (RMSE)

$$\text{RMSE} = \sqrt{\frac{1}{N} \sum_{i=1}^{N} \left(\widehat{S}_i - S_i\right)^2} \tag{14}$$

### Mean Absolute Percentage Error (MAPE)

$$\text{MAPE} = \frac{1}{N} \sum_{i=1}^{N} \frac{\left| \widehat{S}_i - S_i \right|}{S_i} \times 100\% \tag{15}$$

### Coefficient of Determination ($R^2$)

$$R^2 = 1 - \frac{\sum_{i=1}^{N} \left(\widehat{S}_i - S_i\right)^2}{\sum_{i=1}^{N} \left(S_i - \overline{S}\right)^2} \tag{16}$$

where, $N$ is the total number of samples (i.e., cycles), $S_i$ and $\widehat{S}_i$ are the true and computed capacity, respectively. $\overline{S}$ is the sample mean of the ground-truth capacities in the computation horizon.

## Author contributions

**Xinghao Huang:** Writing – review & editing, Writing – original draft, Visualization, Validation, Methodology, Formal analysis, Conceptualization. **Shengyu Tao:** Writing – review & editing, Writing – original draft, Visualization, Validation, Methodology, Formal analysis, Conceptualization. **Chen Liang:** Writing – review & editing, Visualization. **Jiawei Chen:** Writing – review & editing, Methodology, Formal analysis. **Junzhe Shi:** Writing – review & editing, Methodology, Formal analysis. **Bizhong Xia:** Resources, Supervision, Project administration, Funding acquisition, Conceptualization. **Guangmin Zhou:** Supervision, Project administration, Writing – review & editing, Supervision, Conceptualization. **Xuan Zhang:** Supervision, Project administration, Writing – review & editing, Supervision, Conceptualization.

## Declaration of competing interest

The authors declare no known competing interest.

## Acknowledgments

This work was funded by the National Natural Science Foundation of China (Grant No. 51877120).

## Data and code availability

Data used in this work can be available at https://github.com/terencetaothucb/PIMOE-Physics-informed-mixture-of-experts-for-battery-degradation-trajectory-computation.

# Supplementary Information

**Physics-informed mixture of experts (PIMOE) network for interpretable battery degradation trajectory computation amid second-life complexities**

Xinghao Huang [1], Shengyu Tao [1,2,][*], Chen Liang [1], Jiawei Chen [3], Junzhe Shi [2], Yuqi Li [4],
Bizhong Xia [1,][*], Guangmin Zhou[1,][*], Xuan Zhang [1,][*]

[1] Tsinghua Shenzhen International Graduate School, Tsinghua University, Shenzhen, 518055, China
[2] Department of Civil and Environmental Engineering, UC Berkeley, Berkeley, CA, 94720, USA
[3] School of Computer Science, Peking University, Beijing 100871, China
[4] Department of Materials Science and Engineering, Stanford University, Stanford, CA, 94305, USA

* Corresponding authors:
sytao@berkeley.edu (S. Tao); xiabz@sz.tsinghua.edu.cn (B. Xia),
guangminzhou@sz.tsinghua.edu.cn (G. Zhou); xuanzhang@sz.tsinghua.edu.cn (X. Zhang).

This file contains:
Supplementary Figures 1-25;
Supplementary Tables 1-11;
Supplementary Notes 1-11.

**Supplementary Figure 1.** QV curves and relaxation voltage curves of NCA material batteries.

In the UL dataset, the QV curves and relaxation voltage curves of NCA material batteries exhibit a declining trend with aging within the selected input data range. As the SOH decreases, both curves show a downward trend, with different colors mapping distinct SOH values, indicating a strong correlation between the selected input data and battery degradation. (a)-(e) represent batteries with different materials, each randomly selecting one cell for visualization in the corresponding plot.

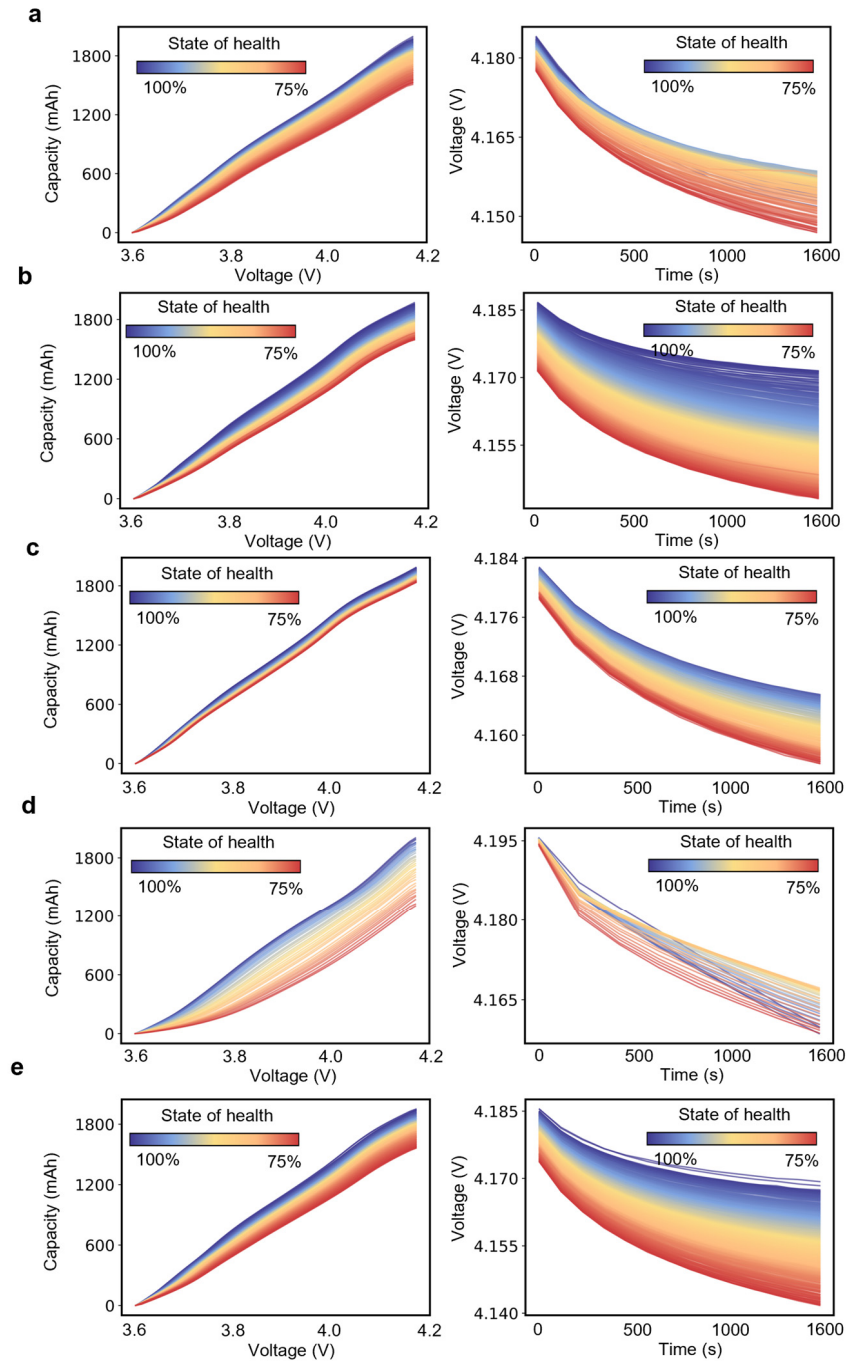

**Supplementary Figure 2.** QV curves and relaxation voltage curves of NCM material batteries.

In the UL dataset, the QV curves and relaxation voltage curves of NCM material batteries exhibit a declining trend with aging within the selected input data range. As the SOH decreases, both curves show a downward trend, with different colors mapping distinct SOH values, indicating a strong correlation between the selected input data and battery degradation. (a)-(e) represent batteries with different materials, each randomly selecting one cell for visualization in the corresponding plot.

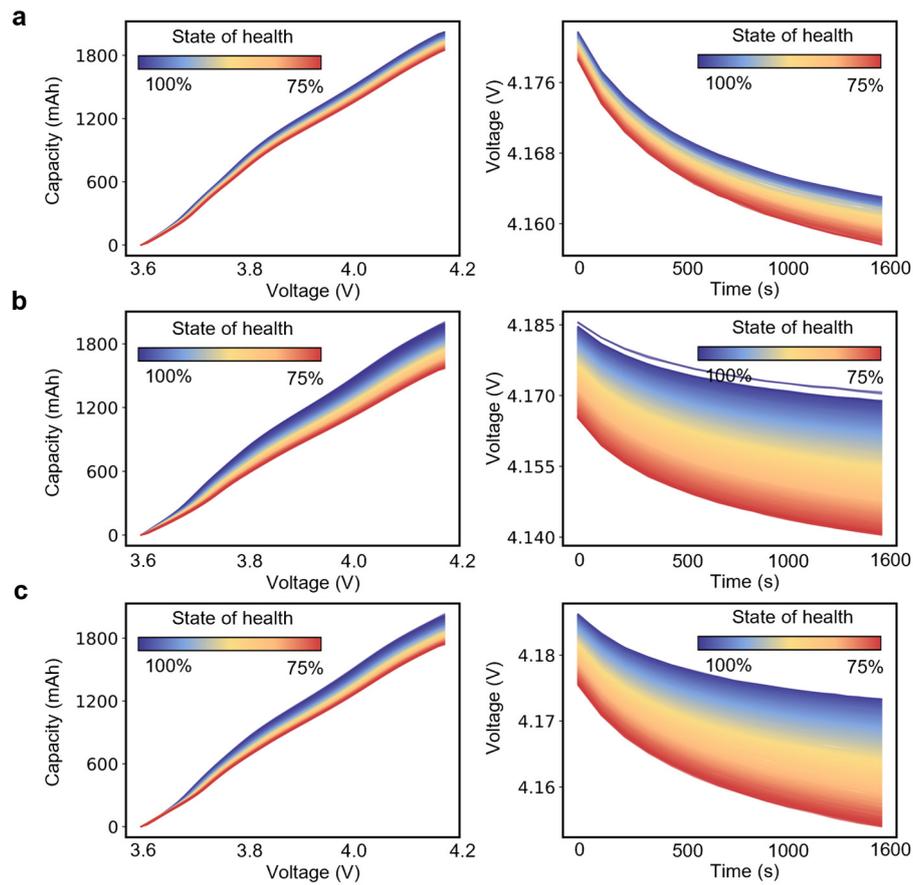

**Supplementary Figure 3.** QV curves and relaxation voltage curves of NCMNCA material batteries.

In the UL dataset, the QV curves and relaxation voltage curves of NCMNCA material batteries exhibit a declining trend with aging within the selected input data range. As the SOH decreases, both curves show a downward trend, with different colors mapping distinct SOH values, indicating a strong correlation between the selected input data and battery degradation. (a)-(e) represent batteries with different materials, each randomly selecting one cell for visualization in the corresponding plot.

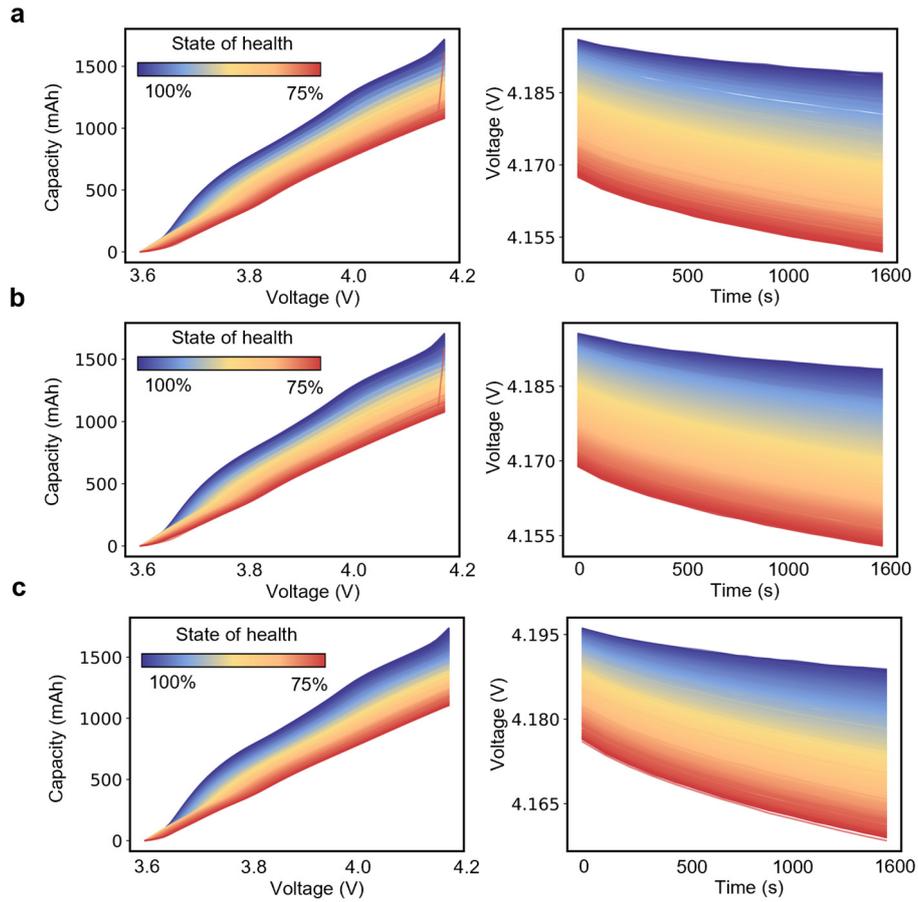

**Supplementary Figure 4.** 12 features in the NCA material batteries.

Aging trends of 12 features in the NCA material battery dataset, with different colors representing different operating conditions. Among these, six features are extracted from the QV curves, and six are derived from the relaxation voltage curves (specific extraction methods are detailed in Supplementary Note 2). The feature values exhibit nearly linear trends over the aging process, demonstrating a strong correlation between these features and battery degradation.

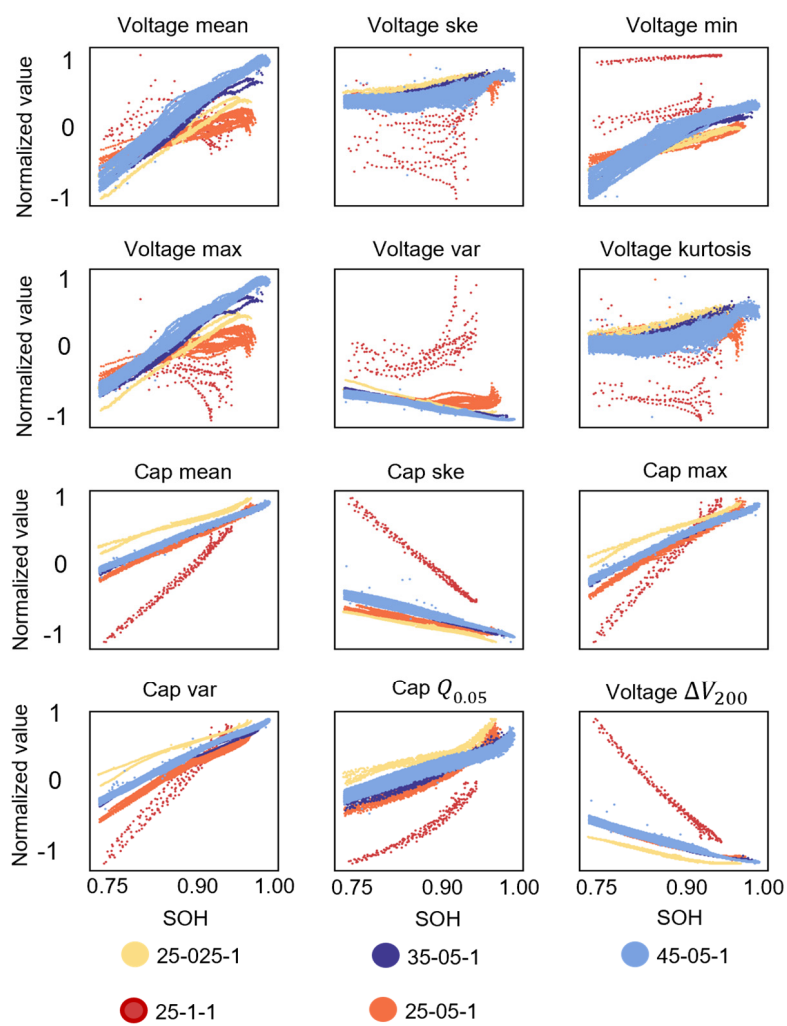

**Supplementary Figure 5.** 12 features in the NCM material batteries.

Aging trends of 12 features in the NCM material battery dataset, with different colors representing different operating conditions. Among these, six features are extracted from the QV curves, and six are derived from the relaxation voltage curves (specific extraction methods are detailed in Supplementary Note 2). The feature values exhibit nearly linear trends over the aging process, demonstrating a strong correlation between these features and battery degradation.

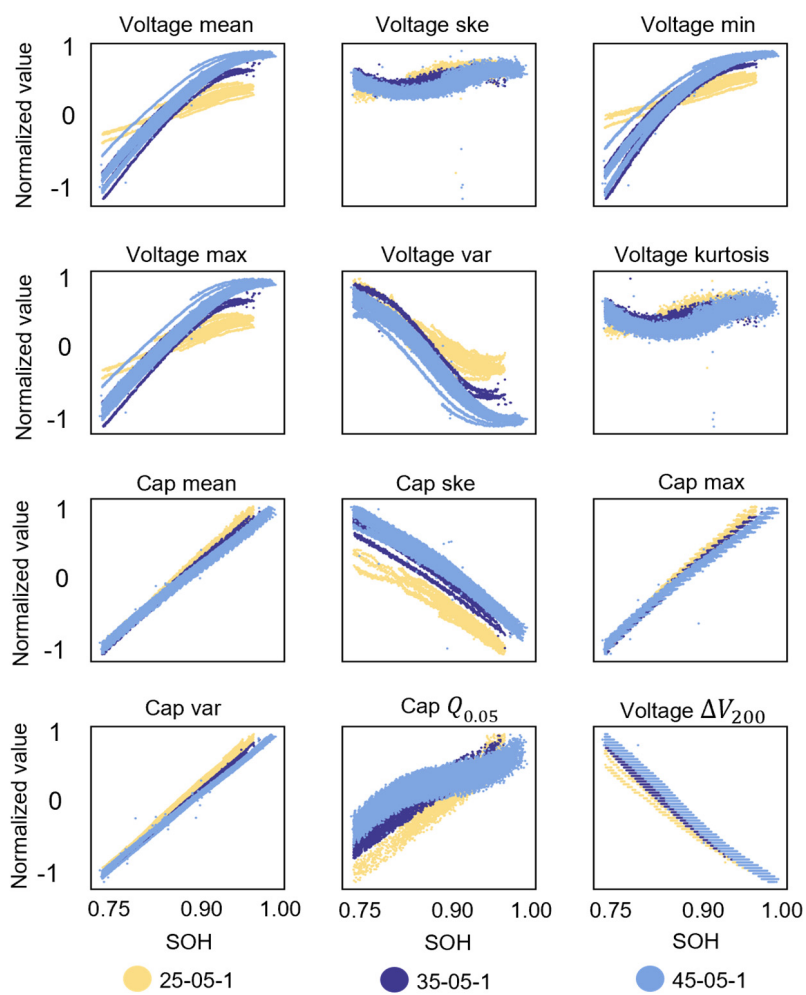

**Supplementary Figure 6.** 12 features in the NCMNCA material batteries.

Aging trends of 12 features in the NCMNCA material battery dataset, with different colors representing different operating conditions. Among these, six features are extracted from the QV curves, and six are derived from the relaxation voltage curves (specific extraction methods are detailed in Supplementary Note 2). The feature values exhibit nearly linear trends over the aging process, demonstrating a strong correlation between these features and battery degradation.

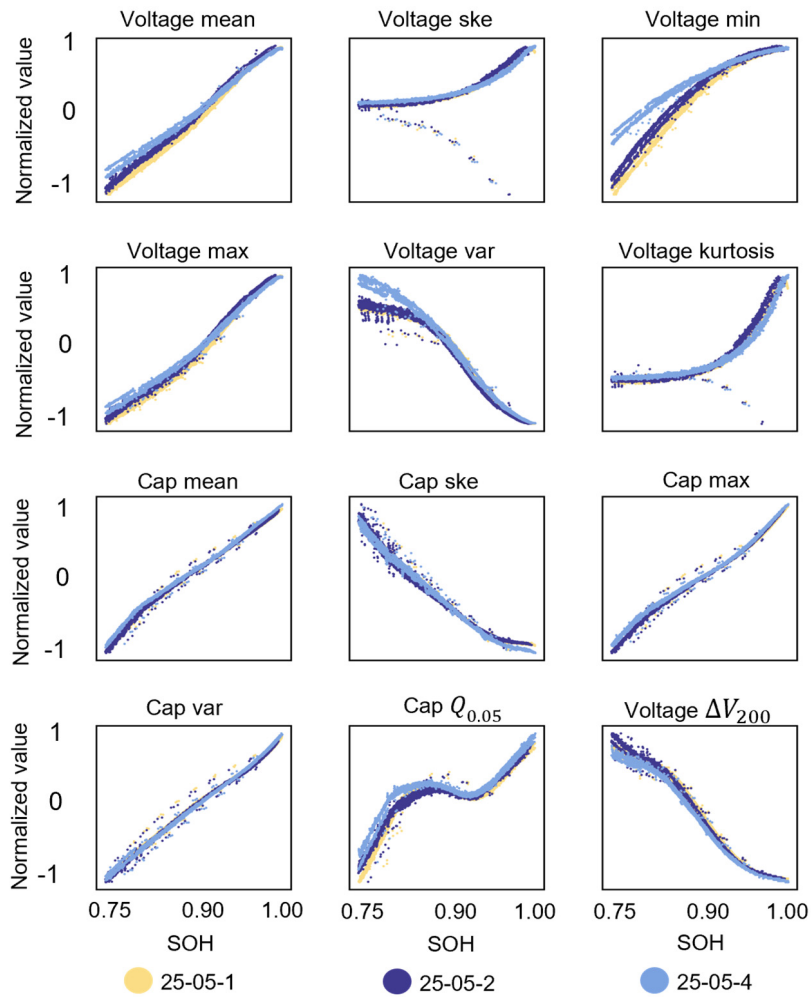

**Supplementary Figure 7.** Machine learning pipeline design of the PIMOE.

The initial input consists of randomly sampled SOC data collected from battery field operations, including capacity-voltage curves and relaxation voltage curves. After feature engineering and standardization, these curves are fed into the degradation router. The router outputs weights corresponding to different expert networks. The original capacity-voltage curves are adaptively resampled, and based on the expert network weights, the predictions from multiple experts are synthesized to output degradation trends. These degradation trends are then cyclically combined with future operating condition predictions to estimate degradation trajectories under uncertain conditions. The loss function incorporates both $Loss_{cv}$ (calculated based on the importance weights of samples from different degradation stages within a batch) and $Loss_{MSE}$ (measuring the discrepancy between predicted and actual values).

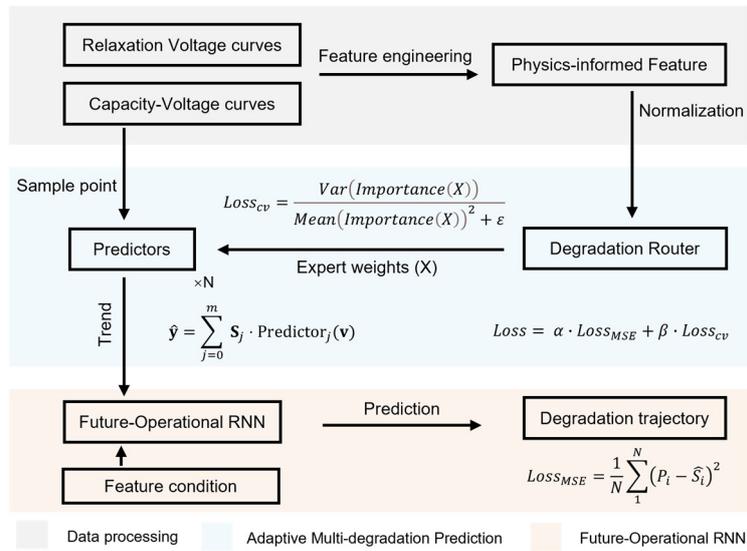

**Supplementary Figure 8.** Prediction Results of Full Lifecycle Health Status for NCA Material Batteries.

Predicting the degradation trajectory of NCA material batteries for arbitrary samples throughout their full lifecycle. A sliding window approach is employed for data extraction and prediction (details provided in Supplementary Note 1). (a) presents the prediction results under the 25-1-1 operating condition, where the typical lifespan is approximately 30 cycles. Accordingly, predictions are made for the next 10 cycles based on the current cycle. (b)-(e) respectively display prediction results under different operating conditions, where only partial data from the current cycle is used to forecast the next 50 cycles for individual batteries. The color gradient represents the health states of different samples. (Naming conventions are specified in Supplementary Table 1).

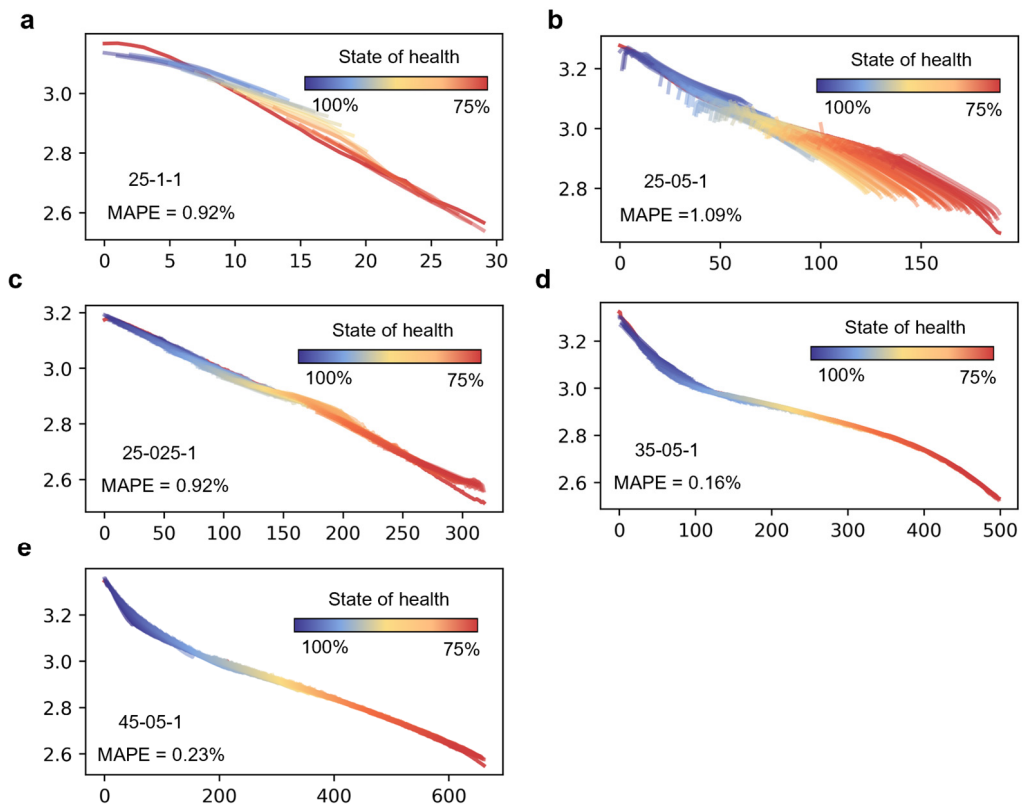

**Supplementary Figure 9.** Prediction Results of Full Lifecycle Health Status for NCM Material Batteries.

Predicting the degradation trajectory of arbitrary NCM material battery samples throughout their full lifecycle. Using a sliding window approach for data extraction and prediction (see Supplementary Note 1 for details), the method predicts the next 50 cycles for individual batteries based solely on partial data from the current cycle. (a)-(c) present prediction results under different operating conditions, with color coding representing the health states of different samples. (Naming conventions are specified in Supplementary Table 1).

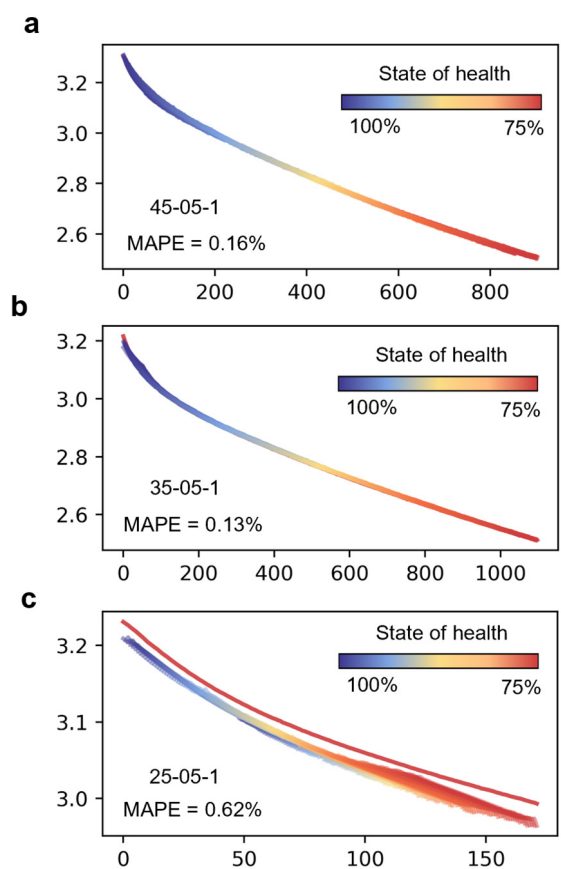

**Supplementary Figure 10.** Prediction Results of Full Lifecycle Health Status for NCMNCA Material Batteries.

Predicting the degradation trajectory of arbitrary NCMNCA material battery samples throughout their full lifecycle. Using a sliding window approach for data extraction and prediction (see Supplementary Note 1 for details), the method predicts the next 50 cycles for individual batteries based solely on partial data from the current cycle. (a)-(c) present prediction results under different operating conditions, with color coding representing the health states of different samples. (Naming conventions are specified in Supplementary Table1).

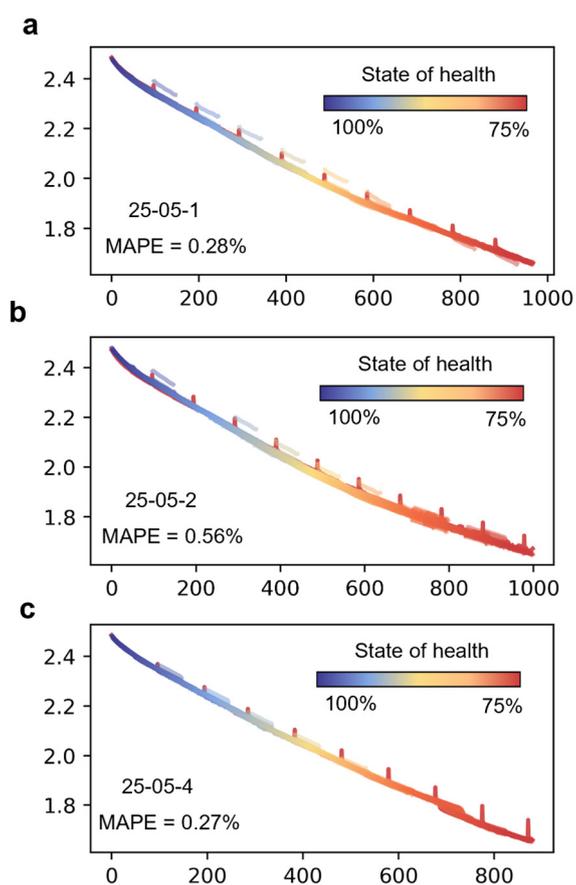

**Supplementary Figure 11.** Prediction Results of Full Lifecycle Health Status for TPSL-Arbitrary load-conditioned batteries.

Predicting degradation trajectories under repurposed load-varying conditions. For TPSL-Arbitrary load-conditioned batteries, the method employs a sliding window approach for data extraction and prediction (detailed in Supplementary Note 1). Using only partial data from the current cycle, it predicts the next 50 cycles for individual batteries, where red points represent predicted values and black points denote the original capacity degradation curves. (a)-(c) display prediction results for different samples, showing PIMOE's accurate prediction of both battery capacity degradation and capacity variations during load condition changes.

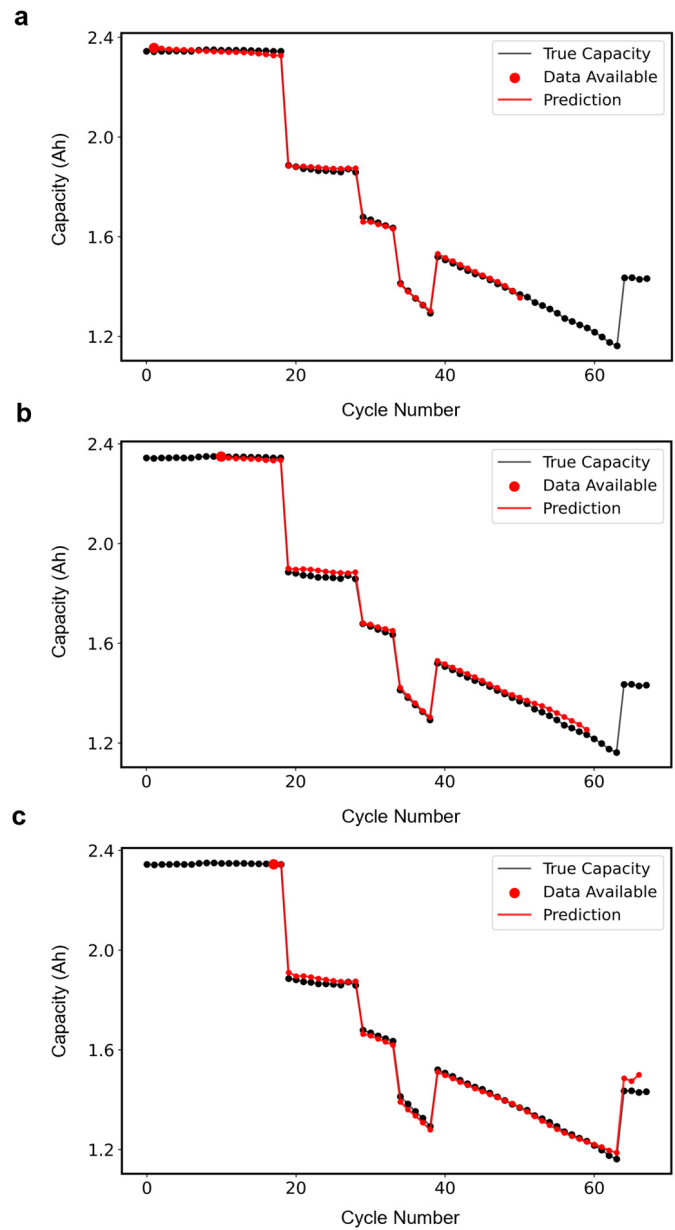

**Supplementary Figure 12.** Prediction Results of Full Lifecycle Health Status for TPSL- Fixed load-conditioned batteries.

Predicting degradation trajectories under repurposed load-varying conditions. For TPSL-Fixed load-conditioned batteries, the method employs a sliding window approach for data extraction and prediction (detailed in Supplementary Note 1). Using only partial data from the current cycle, it predicts the next 50 cycles for individual batteries, where red points represent predicted values and black points denote the original capacity degradation curves. (a)-(c) display prediction results for different samples, showing PIMOE's accurate prediction of both battery capacity degradation and capacity variations during load condition changes.

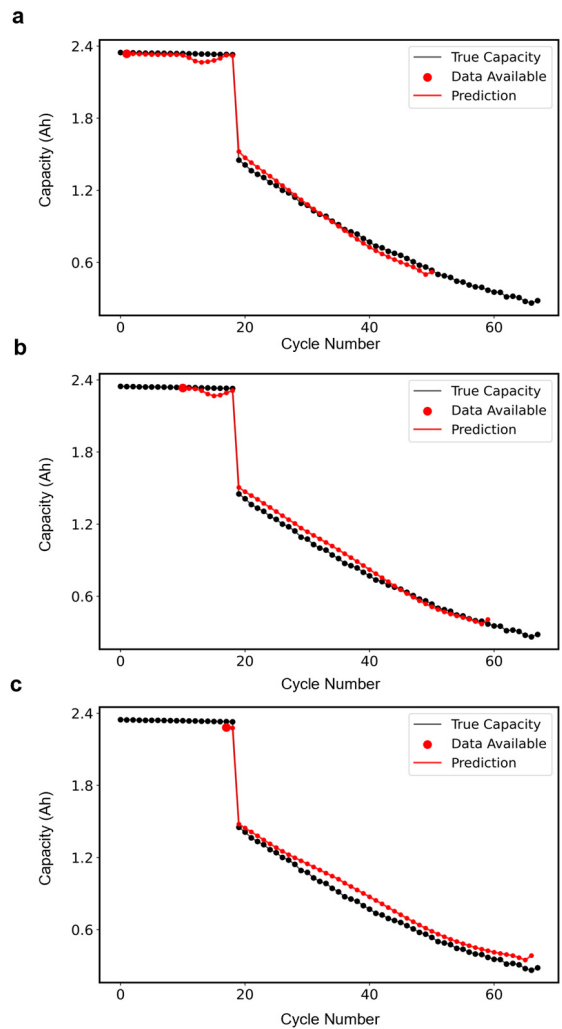

**Supplementary Figure 13.** Prediction Results of Full Lifecycle Health Status for TPSL-Arbitrary load-conditioned batteries.

Predicting degradation trajectories under repurposed load-varying conditions. For TPSL-Fixed load-conditioned batteries, the method employs a sliding window approach for data extraction and prediction (detailed in Supplementary Note 1). Using only partial data from the current cycle, it predicts the next 50 cycles for individual batteries, where red points represent predicted values and black points denote the original capacity degradation curves. (a)-(c) display prediction results for different samples, showing PIMOE's accurate prediction of both battery capacity degradation and capacity variations during load condition changes.

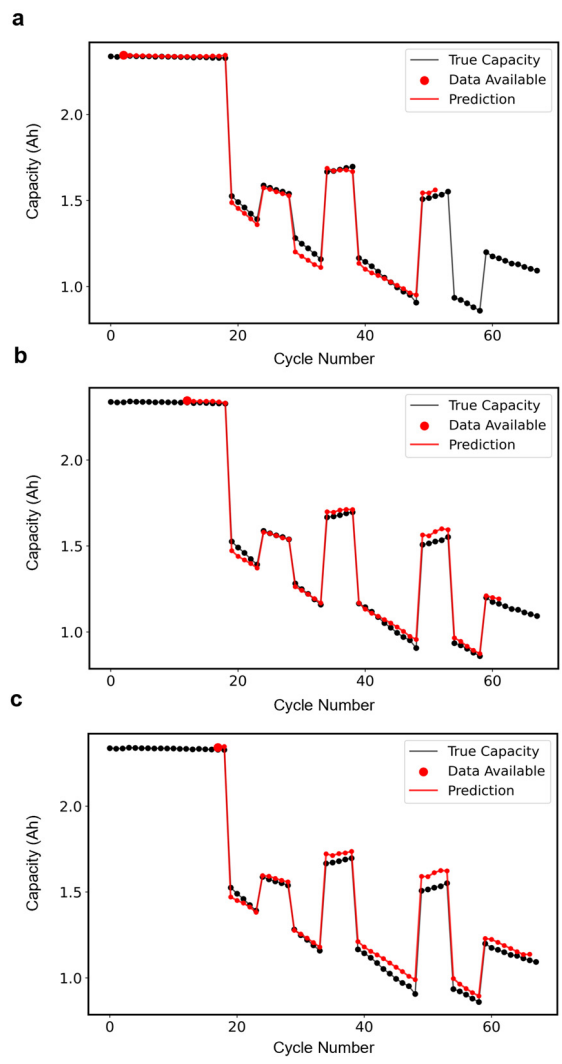

**Supplementary Figure 14.** Prediction Results of Full Lifecycle Health Status for TPSL- Fixed load-conditioned batteries.

Predicting degradation trajectories under repurposed load-varying conditions. For TPSL-Fixed load-conditioned batteries, the method employs a sliding window approach for data extraction and prediction (detailed in Supplementary Note 1). Using only partial data from the current cycle, it predicts the next 50 cycles for individual batteries, where red points represent predicted values and black points denote the original capacity degradation curves. (a)-(c) display prediction results for different samples, showing PIMOE's accurate prediction of both battery capacity degradation and capacity variations during load condition changes.

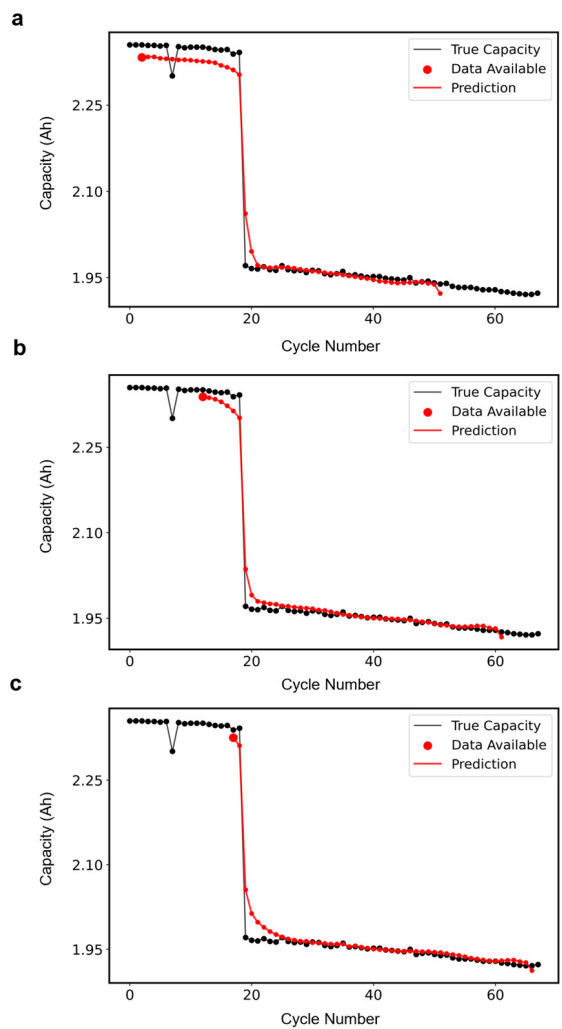

**Supplementary Figure 15.** Correlation between expert weights and degradation stages under different operating conditions.

Correlation between expert weights and degradation stages under different operating conditions. By randomly selecting pre-trained models corresponding to specific conditions and using single-cycle test samples from the full lifecycle of batteries across different datasets as input, we visualize the relationship between the evolving expert weights in the model output and the degradation stages. The horizontal axis represents test samples from different degradation stages, while the vertical axis indicates the weights assigned by five experts to each sample, with color intensity reflecting the magnitude of expert weights. As degradation progresses, different experts dominate corresponding stages, enabling adaptive integration of expert weights to predict degradation trajectories for samples at various degradation phases.

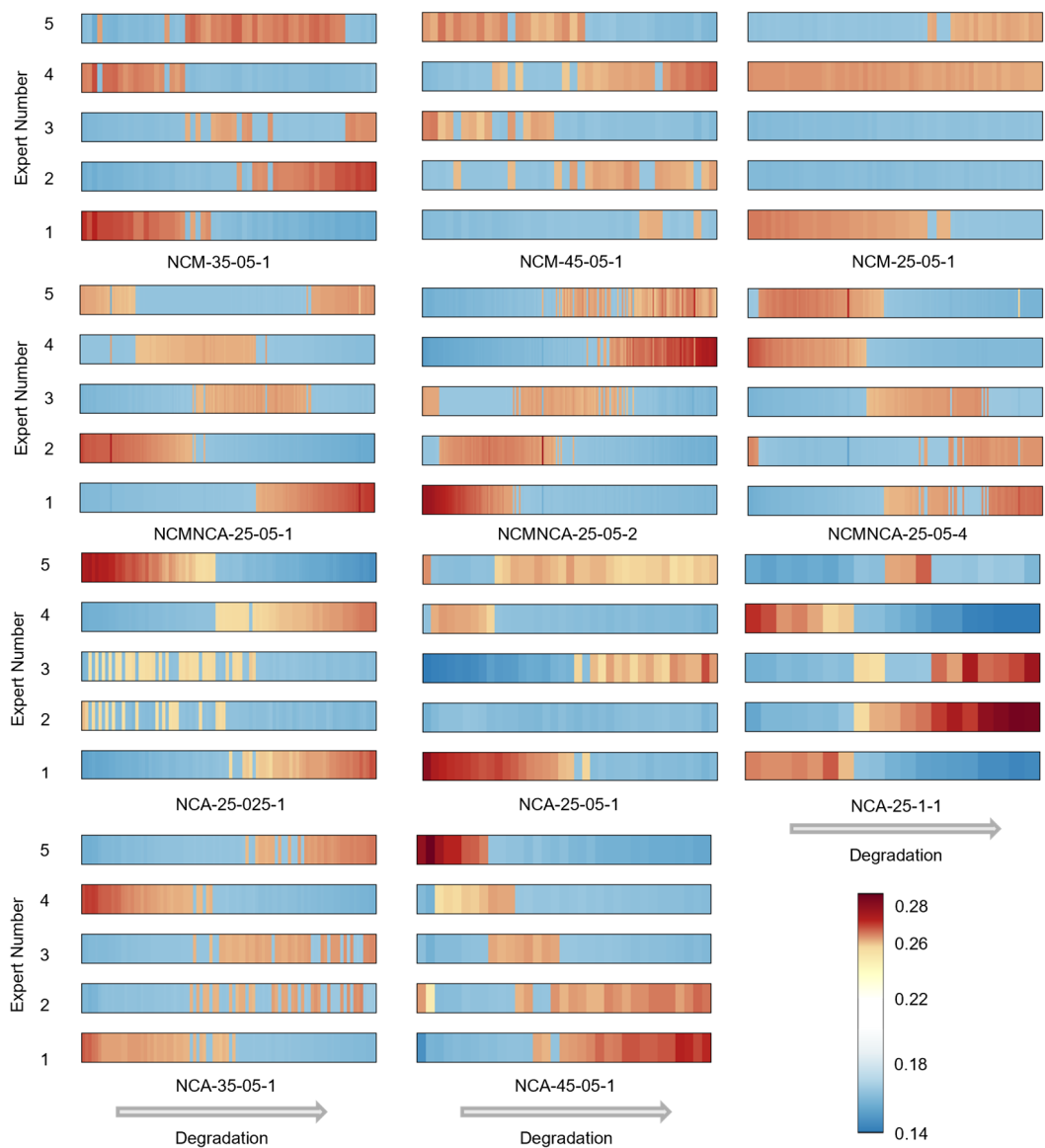

**Supplementary Figure 16.** Expert weights output by the model for an entire batch of NCM-material batteries retired at different SOH levels.

Visualization of expert weights output by the model for an entire batch of batteries retired at different SOH levels during practical deployment. For all NCM-material batteries in selected SOH retirement scenarios, we randomly chose one pre-trained model to visualize the expert weights across the entire battery population. The horizontal axis represents the number of retired batteries, while the vertical axis indicates the weights assigned by five corresponding experts, with color intensity reflecting weight values Fig. (a), (b), and (c) represent three distinct operating conditions: 25-05-1, 35-05-1, and 45-05-1 respectively.

The model enables optimal utilization decisions by processing partial cycle data obtained from retired batteries as input and generating interpretable expert weight outputs. In certain datasets, the number of batteries at 95% SOH differs from those at 75% SOH, as some batteries were retired before reaching 75% SOH, resulting in a reduced test population. Nevertheless, the model achieves highly accurate and interpretable classification results for battery samples at selected SOH levels (see Supplementary Note 7 for details).

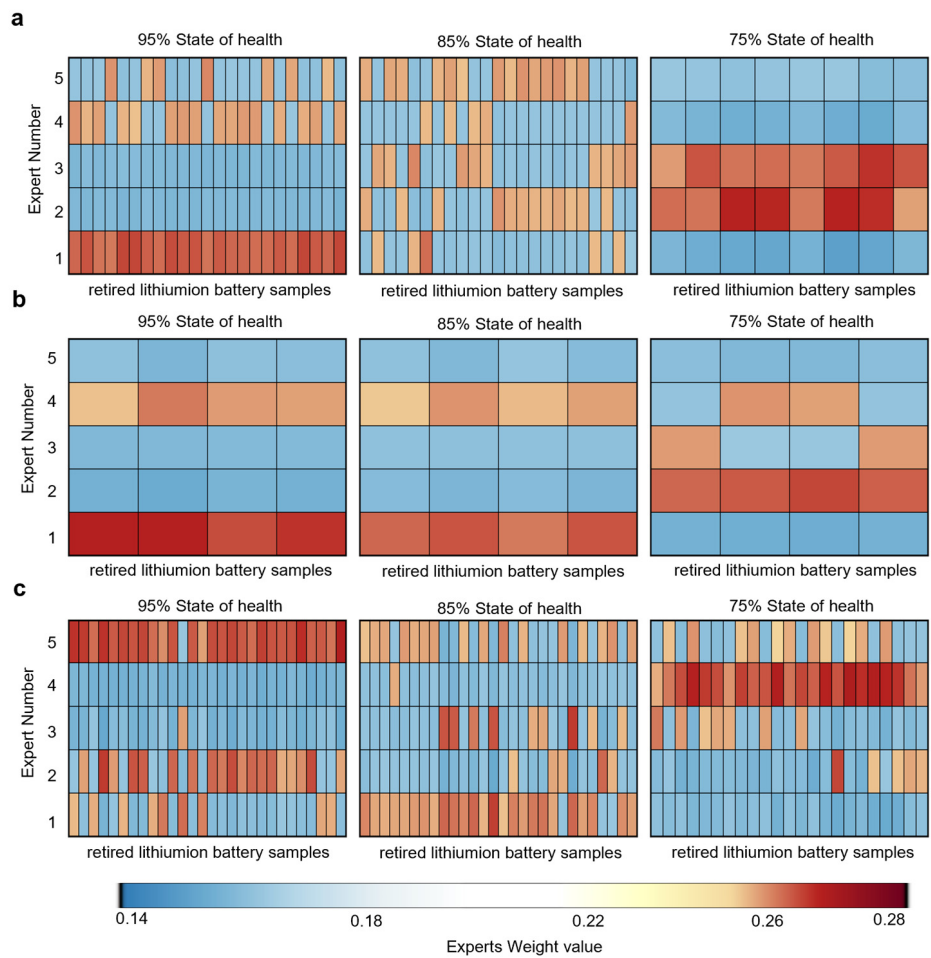

**Supplementary Figure 17.** Expert weights output by the model for an entire batch of NCMNCA-material batteries retired at different SOH levels.

Visualization of expert weights output by the model for an entire batch of batteries retired at different SOH levels during practical deployment. For all NCMNCA-material batteries in selected SOH retirement scenarios, we randomly chose one pre-trained model to visualize the expert weights across the entire battery population. The horizontal axis represents the number of retired batteries, while the vertical axis indicates the weights assigned by five corresponding experts, with color intensity reflecting weight values, Fig. (a), (b), and (c) represent three distinct operating conditions: 25-05-1, 25-05-2, and 25-05-4 respectively.

The model enables optimal utilization decisions by processing partial cycle data obtained from retired batteries as input and generating interpretable expert weight outputs. In certain datasets, the number of batteries at 95% SOH differs from those at 75% SOH, as some batteries were retired before reaching 75% SOH, resulting in a reduced test population. Nevertheless, the model achieves highly accurate and interpretable classification results for battery samples at selected SOH levels (see Supplementary Note 7 for details).

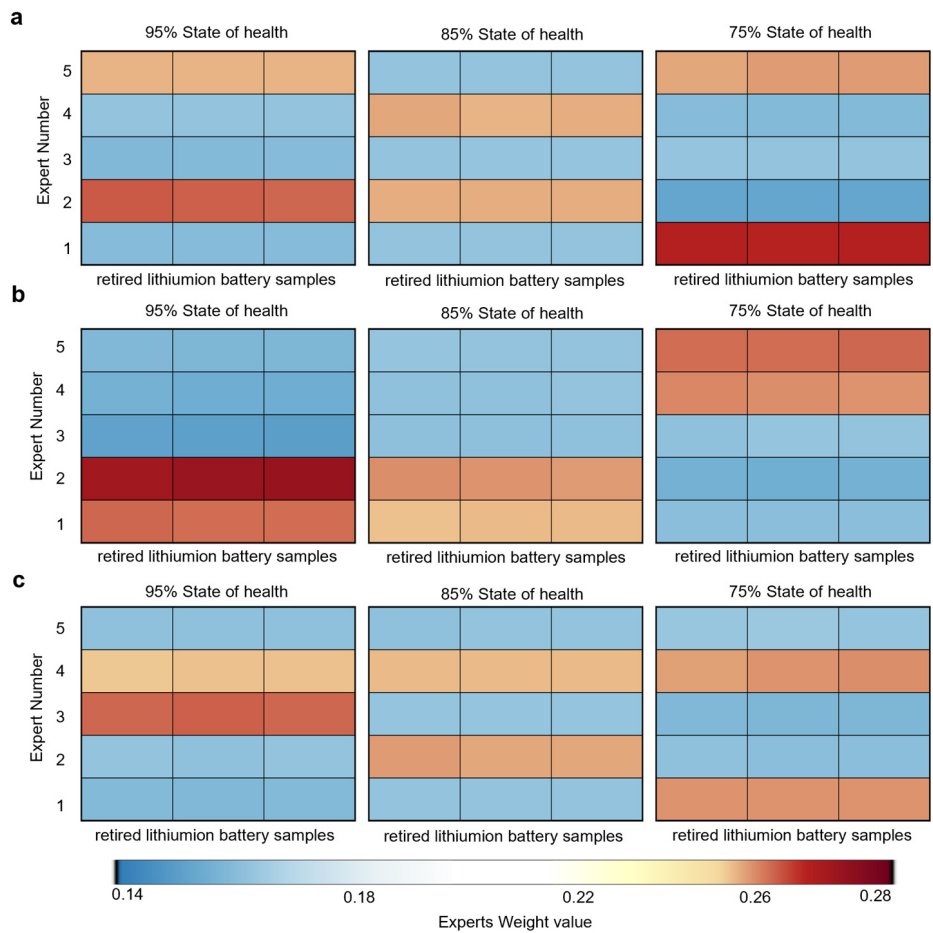

**Supplementary Figure 18.** Expert weights output by the model for an entire batch of NCA-material batteries retired at different SOH levels.

Visualization of expert weights output by the model for an entire batch of batteries retired at different SOH levels during practical deployment. For all NCA-material batteries in selected SOH retirement scenarios, we randomly chose one pre-trained model to visualize the expert weights across the entire battery population. The horizontal axis represents the number of retired batteries, while the vertical axis indicates the weights assigned by five corresponding experts, with color intensity reflecting weight values Fig. (a)-(e) represent three distinct operating conditions: 25-1-1,25-05-1,25-025-1,35-05-1,45-05-1 respectively.

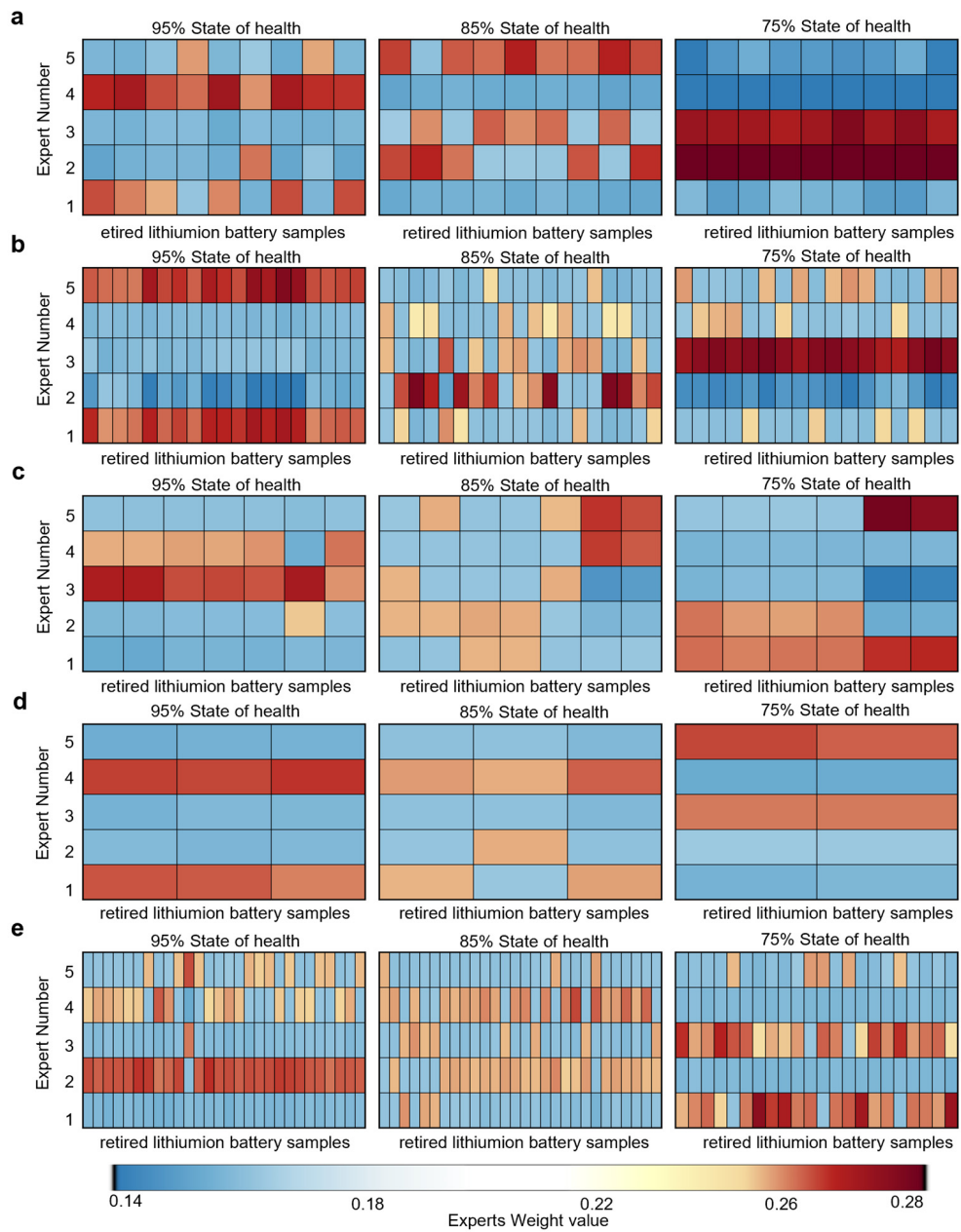

**Supplementary Figure 19.** t-SNE Dimensionality Reduction Visualization of Expert Weights from NCA Battery Test Model.

For NCA-material test batteries under different operating conditions in the UL dataset, we randomly selected a pre-trained model and used full-lifecycle single-cycle test samples as input. The expert weights output by the model were visualized after t-SNE clustering, with samples from different aging stages color-mapped accordingly (see Supplementary Note 8 for details). The results show clear separation between healthy and aged batteries in the two-dimensional latent space after clustering, further demonstrating the effectiveness of using expert weights for second-life utilization decisions. Fig. (a)-(d) represents three distinct operating conditions.

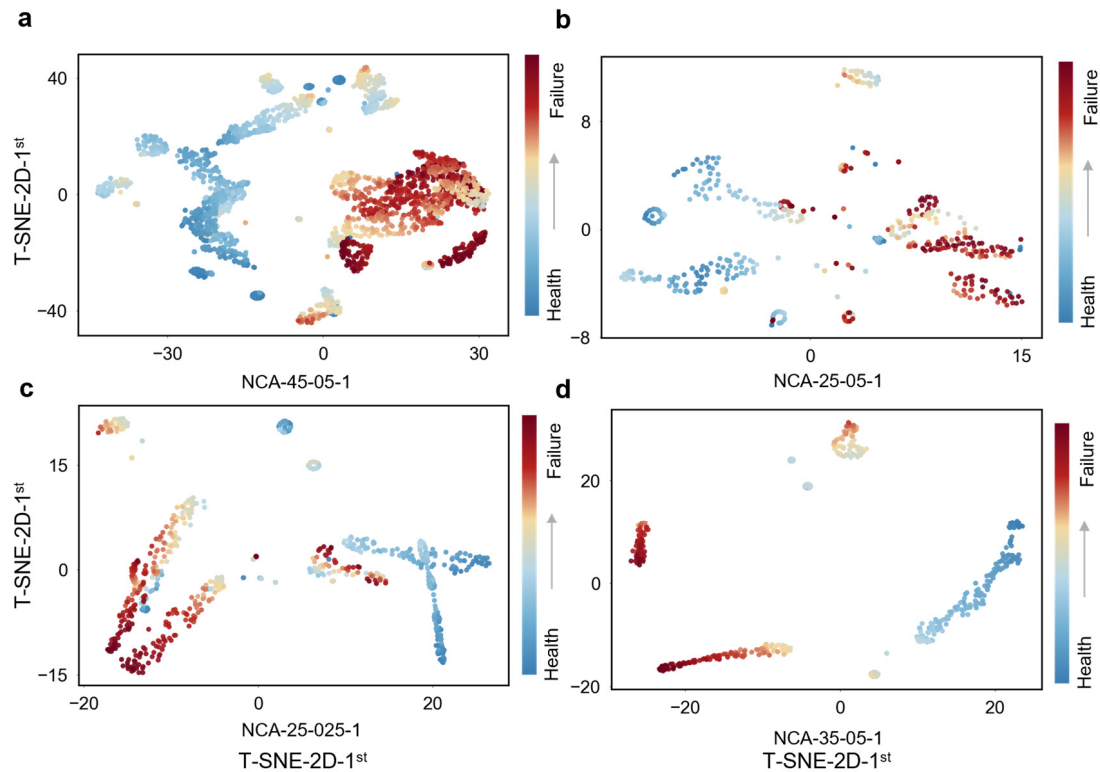

**Supplementary Figure 20.** t-SNE Dimensionality Reduction Visualization of Expert Weights from NCMNCA Battery Test Model.

For NCMNCA-material test batteries under different operating conditions in the UL dataset, we randomly selected a pre-trained model and used full-lifecycle single-cycle test samples as input. The expert weights output by the model were visualized after t-SNE clustering (see Supplementary Note 8 for details), with samples from different aging stages color-mapped accordingly. The results show clear separation between healthy and aged batteries in the two-dimensional latent space after clustering, further demonstrating the effectiveness of using expert weights for second-life utilization decisions. Fig. (a)-(c) represents three distinct operating conditions.

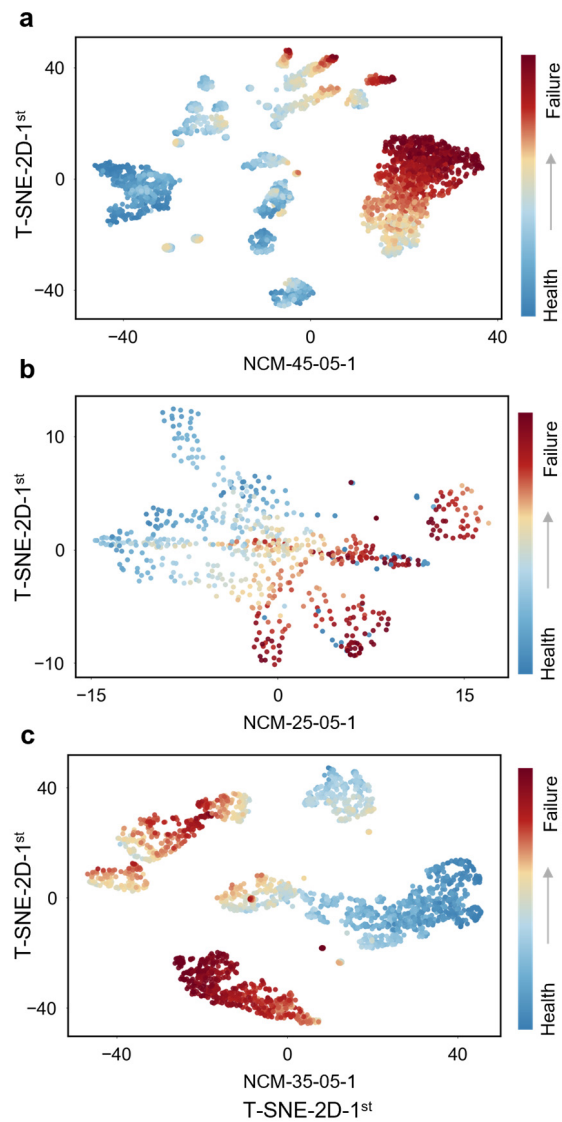

**Supplementary Figure 21.** t-SNE Dimensionality Reduction Visualization of Expert Weights from NCM Battery Test Model.

For NCM-material test batteries under different operating conditions in the UL dataset, we randomly selected a pre-trained model and used full-lifecycle single-cycle test samples as input. The expert weights output by the model were visualized after t-SNE clustering (see Supplementary Note 8 for details), with samples from different aging stages color-mapped accordingly. The results show clear separation between healthy and aged batteries in the two-dimensional latent space after clustering, further demonstrating the effectiveness of using expert weights for second-life utilization decisions. Fig. (a)-(c) represents three distinct operating conditions.

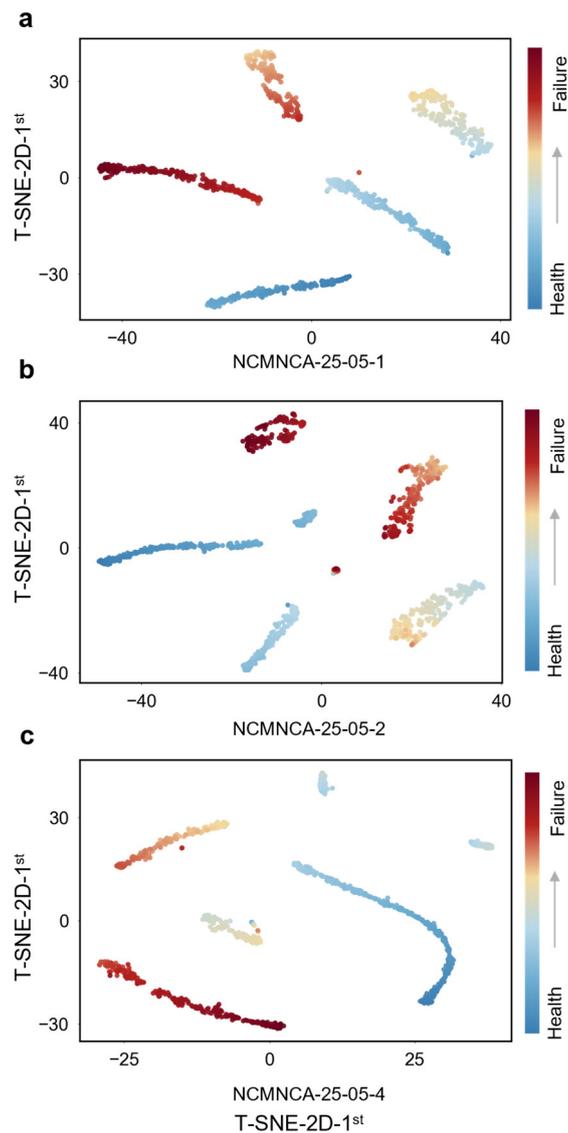

**Supplementary Figure 22.** Sensitivity Analysis of Physics-informed Feature Noise Across Batteries with Divergent Degradation Paths.

Analysis of Feature Impact on Model Performance Across Batteries with Divergent Degradation Paths To investigate feature importance, we introduced random Gaussian noise to each extracted feature from raw data, thereby influencing the weight generation for degradation routing in the PIMOE model. Fig.(a) demonstrates the performance variations of all 12 features under noise perturbation, with the red baseline indicating the original noise-free model performance. Each experiment was repeated 10 times to ensure reliability.

Fig.(b) displays the relationship between predicted and true values across full lifecycle samples under normal conditions, while Fig.(c) shows this relationship for the noise-sensitive feature (Cap mean). The results reveal significant performance degradation when samples were incorrectly assigned to expert networks, further validating the effectiveness of PIMOE's two-stage approach that first distinguishes aging phases before predicting degradation trajectories.

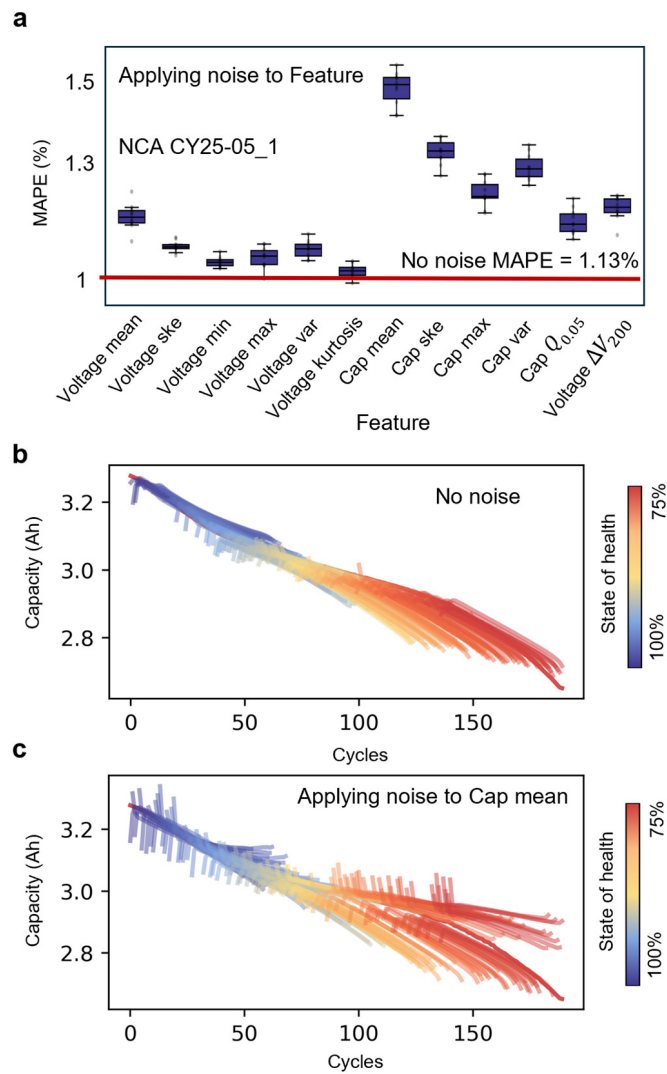

**Supplementary Figure 23.** Sensitivity Analysis of Physics-informed Feature Noise Across Batteries with Similar Degradation Paths.

Analysis of Feature Impact on Model Performance Across Batteries with Similar Degradation Paths To investigate feature importance, we introduced random Gaussian noise to each extracted feature from raw data, thereby influencing the weight generation for degradation routing in the PIMOE model. Fig.(a) demonstrates the performance variations of all 12 features under noise perturbation, with the red baseline indicating the original noise-free model performance. Each experiment was repeated 10 times to ensure reliability.

Fig.(b) displays the relationship between predicted and true values across full lifecycle samples under normal conditions, while Fig.(c) shows this relationship for the noise-sensitive feature (Voltage max). The results show that even when samples were incorrectly assigned to expert networks, the model performance did not degrade significantly. This observation may be attributed to the inherently small differences in degradation paths among these aged batteries, where different experts performed similar prediction tasks.

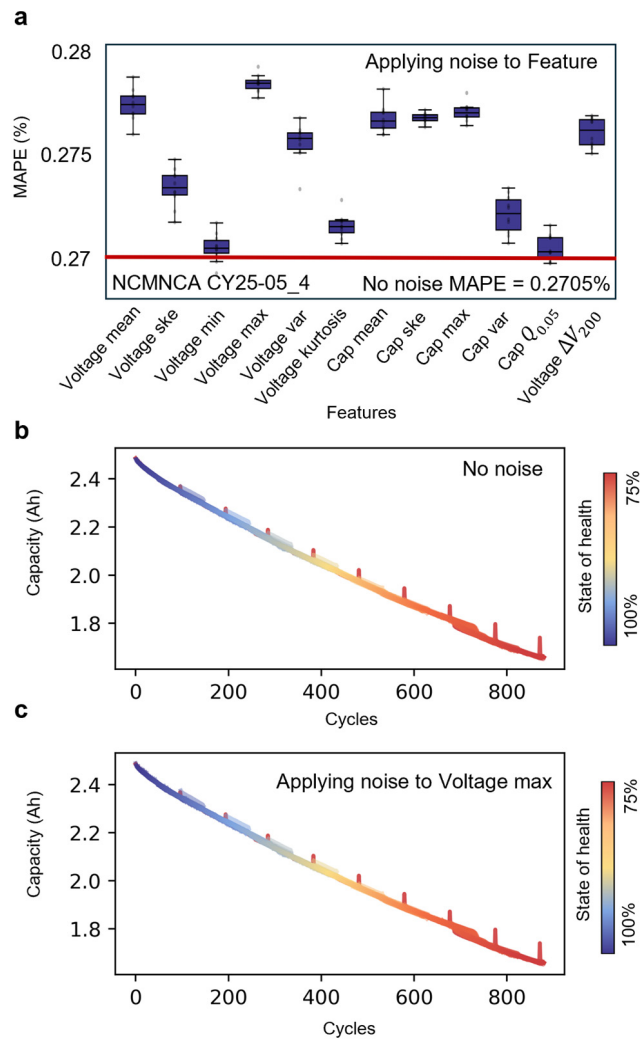

**Supplementary Figure 24.** Impact of Different Physics-informed Feature on Model Performance.

Here we present the performance changes of the model when reducing physics-informed features. "Both" denotes the complete model incorporating both capacity-voltage curves and relaxation-voltage curves. "no QV features" indicates the removal of six features extracted from capacity-voltage curves, while "no RV features" represents the exclusion of six features derived from relaxation-voltage curves. The results show that removing RV features leads to relatively greater performance degradation, suggesting that relaxation-voltage features may more effectively guide the degradation routing. Nevertheless, the model maintains relatively stable performance overall, demonstrating that the proposed PIMOE does not depend on any single specific feature to achieve good performance, though multi-dimensional physics-informed features generally yield better results.

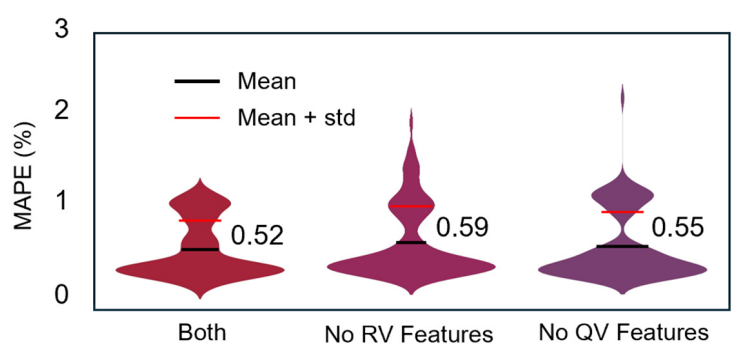

**Supplementary Figure 25.** Performance Comparison of the Proposed PIMOE Model With and Without Historical Data.

We further investigate the applicability and performance of the PIMOE model when directly predicting future degradation trajectories using existing maximum capacity data. Notably, since no within-cycle data is involved, we employ historical capacity to drive the weight allocation in our degradation router, treating capacity magnitude and degradation trajectory as classification criteria for expert network assignment. Each expert network directly outputs degradation trends based on historical cycles rather than capacity-voltage curves.

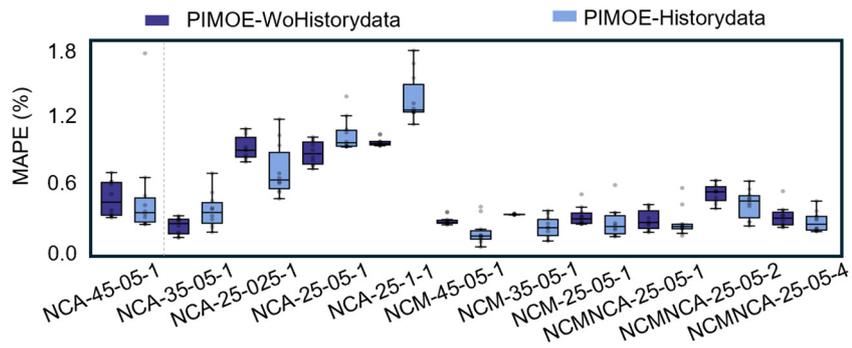

**Supplementary Note 1.** Data preprocessing and extraction procedures.

Here we show the data preprocessing and extraction procedures employed in our experimental study. Given the practical operating conditions, the presence of missing values and sampling errors in raw data represents a common scenario [1]. It should be emphasized that this research deliberately refrains from implementing sophisticated data preprocessing techniques to preserve the practical accessibility of field data, notwithstanding the potential compromise in model performance due to inherent and unavoidable measurement errors [2,3].

The data cleaning protocol involved the systematic removal of anomalous cycles meeting any of the following criteria: cycles exhibiting maximum capacity deviations exceeding 200mAh when compared to both preceding and subsequent cycles, cycles demonstrating zero capacity change during charging phases, and cycles showing abnormal current fluctuations during constant-current charging processes - all of which were identified as characteristic indicators of sampling irregularities. The final processed dataset was constructed by retaining only the continuous cycles that passed these quality control measures.

Let $Q_t \in \mathbb{R}^d$, represent the charging curve of the battery at cycle t, and $F_t \in \mathbb{R}^{12}$ denote the physical information features at cycle t, where d is the number of sampling points. The model predicts the maximum discharge capacity sequence for the next H cycles (see Supplementary Material X for specific prediction horizon). The load conditions for the next H cycles after cycle t are given by $C_t \in \mathbb{R}^{H \times 4}$.

Sample pairs $(X, Y)$ are constructed using a sliding window approach:

$$\mathbf{X} = \begin{bmatrix} (Q_t, F_t, C_t) \\ (Q_{t+1}, F_{t+1}, C_{t+1}) \\ \vdots \\ (Q_{t+N-1}, F_{t+N-1}, C_{t+N-1}) \end{bmatrix} \quad (1)$$

$$\mathbf{Y} = \begin{bmatrix} y_{t+1} & y_{t+2} & \cdots & y_{t+H} \\ y_{t+2} & y_{t+3} & \cdots & y_{t+H+1} \\ \vdots & \vdots & \ddots & \vdots \\ y_{t+N} & y_{t+N+1} & \cdots & y_{t+N+H-1} \end{bmatrix} \quad (2)$$

where N represents the maximum number of cycles for the battery.

**Supplementary Note 2.** Feature extraction.

Inspired by Zhu et al., we extracted six statistical features from the relaxation voltage curve[4] (Max, Mean, Min, Var, Ske, Kur) and four from the charging capacity curve (Max, Mean, Var, Kur), with mathematical formulations detailed in Equation (3)-(8). Recognizing the strong correlation between increasing internal resistance and battery degradation during aging, we specifically designed two key physical-informative features to quantify lithium-ion activity loss: (1) $Q_{0.05}$, representing the charged capacity during a 0.05V voltage rise from the sampled SOC point; and (2) $\Delta V_{200}$, denoting the voltage change corresponding to a 200mAh capacity increase from the same SOC reference. These dual features capture distinct degradation mechanism-$Q_{0.05}$ reflects charge acceptance under mild polarization[5,6], while $\Delta V_{200}$ characterizes voltage hysteresis caused by capacity fade and electrode deterioration[7–9].

The twelve physical-informative features (six from each curve type) were normalized to F ∈ [0,1] for scale alignment, ensuring training stability. It should be noted that the input for driving the degradation gating router may include, but is not limited to, the features mentioned (e.g., IC curve peaks[10–12]). For the TPSL dataset, we exclusively used charging curve-derived features (excluding relaxation voltage characteristics) to demonstrate methodological generality. The standardization formula is as follows:

$$Maxima\ (Max): x_{max} = max\{x_i\} \qquad (3)$$

$$Mean\ (Mean): \bar{x} = \frac{1}{n}\sum_{i=1}^{n} x_i \qquad (4)$$

$$Minima\ (Min): min\{x_i\} \qquad (5)$$

$$Variance\ (Var): \mu_2 = \frac{1}{n-1}\sum_{i=1}^{n}(x_i - \bar{x})^2 \qquad (6)$$

$$Skewness\ (Ske): \mu_3 = \frac{1}{n}\sum_{i=1}^{n}\left(\frac{x-\bar{x}}{\sqrt{\mu_2}}\right)^3 \qquad (7)$$

$$Excess\ Kurtosis\ (Kur): \mu_4 = \frac{1}{n}\sum_{i=1}^{n}\left(\frac{x-\bar{x}}{\sqrt{\mu_2}}\right)^4 - 3 \qquad (8)$$

**Supplementary Note 3.** Selection of prediction horizons.

In practical applications, while extending the prediction horizon of degradation trajectories enhances model utility, excessive prediction spans may compromise accuracy as illustrated in Fig. 5d. Nevertheless, our proposed solution maintains robust performance even when predicting 150 cycles. To balance prediction accuracy with engineering practicality, this study adopts a 50-cycle prediction horizon as the benchmark (representing approximately 30% of a typical battery's median lifespan). Notably, for special operating conditions like NCA-25-1-1 with shorter cycle lives (~30 cycles), the prediction length is correspondingly reduced to 10 cycles to ensure reliability.

It should be emphasized that while existing literature reports longer prediction horizons [13], such methods typically require fixed sampling intervals and complete historical cycle data as input, a design that limits their deployment in dynamic operating conditions [14]. In contrast, our innovative approach requires only partial current cycle data as input and adapts to random initial SOC sampling strategies. This breakthrough not only significantly enhances real-world applicability but also provides technical assurance for end-user deployment convenience.

This design philosophy aligns perfectly with industry demands for "plug-and-play" prediction models, achieving an optimal compromise between prediction accuracy and data acquisition simplicity.

**Supplementary Note 4.** Implementation details of baselines and PIMOE.

PatchTST [15] is a Transformer-based model utilizing patching technique. It enables effective pre-training and transfer learning across datasets. The source code is available at (https: //github.com/yuqinie98/PatchTST).

Informer [16] is a Transformer-based model specifically designed for long-sequence time-series forecasting. It introduces a ProbSparse self-attention mechanism to improve computational efficiency while maintaining prediction accuracy. The model also features a memory-efficient architecture that enables effective handling of large-scale datasets. The official implementation is available at (https://github.com/zhouhaoyi/Informer2020).

Both PatchTST and Informer were implemented according to their official source codes. Where applicable, the baseline models and our proposed method used the same hyperparameters. We tuned the number of layers with reference to [13,17], the neuron number in hidden linear layers from [18], the weight decay in Adam optimizer within [0, 0.0001], and the number of attention heads for multi-head attention from [19,20]. The patch size was set to 3 based on the Q-V curve length[15]. The batch size was set to [8, 16, 64, 128]. We applied for a learning rate of 0.001 across all datasets and selected the best-performing model on the validation set as our final model.

**Supplementary Note 6.** Explanation map.

We define a vector Trend to represent the output of the AMDP module and visualize the hidden outputs of input samples across the full lifecycle to explain the performance improvement of PIMOE. In Fig.4e, using the NCA-35-05-1 battery, we normalize each value of the Trend vector across different samples for clearer interpretation, thereby demonstrating how the model internally processes samples at different aging stages and clarifying its contribution to degradation trajectory prediction. For the input X of the PIMOE model, the Trend is obtained through the AMDP module.

$$\text{Trend} = \text{AMDP}(X) \tag{9}$$

Trend $\in \mathbb{R}^{n \times L}$, where n denotes the number of samples and L represents the prediction length.

**Supplementary Note 7.** Deployment strategy for utilizing pre-trained models to make direct utilization decisions.

  Here we demonstrate the deployment strategy for utilizing pre-trained models to make direct reuse decisions on retired batteries, as samples from different aging stages exhibit distinct differentiation in expert network weights. The trained model enables battery recyclers to predict degradation trajectories under uncertain operating conditions while making cost-effective utilization decisions, as the expert weights show strong correlation with degradation trends through PIMOE's latent subspace clustering where different experts dominate prediction at respective stages. We implement a simple three-path classification system (high-efficiency secondary use for demanding applications like energy storage when Expert 1 dominates, direct recycling when Expert 5 dominates late degradation, and low-efficiency use for electronics in intermediate cases) based on single-cycle field data and future load condition inputs. The effectiveness of this expert weight-based classification is demonstrated in Fig. 4d, providing real-time decision-making with minimal data requirements while enabling adaptive resource allocation and cost optimization through precise battery health grading for practical large-scale retirement scenarios.

**Supplementary Note 8.** t-distributed stochastic neighbor embedding method.

The t-distributed stochastic neighbor embedding (t-SNE) method was employed to perform nonlinear dimensionality reduction on high-dimensional battery degradation features, enabling visualization of their low-dimensional distributions to reveal the relationship between degradation patterns at different cycle stages and expert weights. For the input feature matrix $X \in \mathbb{R}^{n \times d}$ (where n represents the number of samples and d=5 corresponds to the number of experts), pairwise conditional probabilities between samples were calculated to construct the low-dimensional embedding that preserves these relationships in a visually interpretable space. This approach effectively captures the intrinsic clustering structure of battery degradation states while maintaining the relative associations between samples' expert weight distributions across various aging phases

$$p_{j|i} = \frac{\exp(-\|\mathbf{x}_i - \mathbf{x}_j\|^2 / 2\sigma_i^2)}{\sum_{k \neq i} \exp(-\|\mathbf{x}_i - \mathbf{x}_k\|^2 / 2\sigma_i^2)} \quad (10)$$

where $\sigma_i$ is adaptively determined through perplexity (set to 30). By minimizing the KL divergence between the high-dimensional and low-dimensional distributions, the t-SNE algorithm preserves the local structure among samples in the original high-dimensional feature space while transforming global topological relationships into intuitively visible 2D distribution patterns, thereby clearly revealing the correlation characteristics between expert weights and battery degradation states.

$$\mathcal{L} = \sum_i KL(P_i \| Q_i) = \sum_i \sum_j p_{j|i} \log \frac{p_{j|i}}{q_{j|i}} \quad (11)$$

The similarity $q_{j|i}$ in the low-dimensional space is computed using a t-distribution:

$$q_{j|i} = \frac{(1+\|\mathbf{y}_i - \mathbf{y}_j\|^2)^{-1}}{\sum_{k \neq l}(1+\|\mathbf{y}_k - \mathbf{y}_l\|^2)^{-1}} \quad (12)$$

The t-SNE visualization plots for different datasets are provided in Supplementary Fig.18-20.

**Supplementary Note 9.** Degradation Trajectory Prediction Method (Applicable to Scenarios with Complete Historical Data and Consistent Operating Conditions).

The experiments in this section focus on predicting the future degradation trajectory of retired batteries under the condition that historical data is available and future operating conditions remain unchanged. However, this scenario imposes stringent requirements on historical data collection and future usage predictions, which often deviate from real-world applications [21]. To demonstrate that relatively simple methods can achieve satisfactory prediction results, we employ polynomial fitting and a three-layer MLP network. For the polynomial fitting method, given a total cycle count N for the battery, we define a sliding window of size w with a step size s. For the t-th window:

$$\mathcal{W}_t = \{y_t, y_{t+1}, \ldots, y_{t+\omega-1}\}, t \in [0, N - \omega - s] \tag{13}$$

where $y_i$ represents the capacity at the i-th cycle. For each sliding window, we construct a polynomial regression model to capture the capacity degradation trend within that specific segment of the battery's lifespan. This approach models the relationship between cycle number and capacity using a polynomial function, allowing for flexible fitting of nonlinear degradation patterns observed in different phases of the battery's aging process.

$$\hat{y}(x) = \beta_0 + \beta_1 x + \beta_2 x^2 + \cdots + \beta_d x^d \tag{14}$$

where $x \in [t, t + \omega - 1]$ represents cycle index, the model parameters $\hat{\boldsymbol{\beta}} \in \mathbb{R}^{d+1}$ are determined via least squares:

$$\hat{\boldsymbol{\beta}} = \arg\min_{\boldsymbol{\beta}} \sum_{i=t}^{t+\omega-1} (y_i - \hat{y}(i))^2 \tag{15}$$

the fitted model is used to predict the next S cycles:

$$\hat{y}_{t+\omega+k} = \hat{\beta}_0 + \hat{\beta}_1(t + \omega + k) + \cdots + \hat{\beta}_d(t + \omega + k)^d, k = 0, \ldots, s - 1 \tag{16}$$

We employ a cubic polynomial (d=3) for fitting, with a window size ω=50 cycles, prediction step s=50, and sliding window stride of 1. For the MLP, we adopt an end-to-end processing method that models the prediction of future degradation trajectories from historical capacity data as a time series forecasting problem. Given a capacity sequence $Q = \{q_1, q_2, \ldots, q_T\}$, with defined window size w and prediction step s, the k-th sample pair is expressed as:

$$\begin{cases} \mathbf{X}_k = [q_{k\delta}, q_{k\delta+1}, \ldots, q_{k\delta+w-1}] \\ \mathbf{Y}_k = [q_{k\delta+w}, q_{k\delta+w+1}, \ldots, q_{k\delta+w+s-1}] \end{cases} \tag{17}$$

Where $k \in \left[0, \left\lfloor \frac{T-w-s}{\delta} \right\rfloor\right]$. A three-layer MLP network is designed:

$$\begin{aligned} \mathbf{h}_1 &= \text{ReLU}(\mathbf{W}_1 \mathbf{X} + \mathbf{b}_1) \\ \mathbf{h}_2 &= \text{ReLU}(\mathbf{W}_2 \mathbf{h}_1 + \mathbf{b}_2) \\ \hat{Y} &= \mathbf{W}_3 \mathbf{h}_2 + \mathbf{b}_3 \end{aligned} \tag{18}$$

Where weight matrices $\mathbf{W}_1 \in \mathbb{R}^{128 \times w}, \mathbf{W}_2 \in \mathbb{R}^{64 \times 128}, \mathbf{W}_3 \in \mathbb{R}^{s \times 64}$. In our experiments, we set the window size $\omega = 10$ cycles, prediction step s=50, and sliding window stride to 1.

**Supplementary Note 10.** Data reduction strategy.

The availability of retired battery data significantly hinders the transition of data-driven approaches from laboratory to real-world deployment [21–23]. This section demonstrates how to evaluate the robustness of the proposed model under data-constrained conditions. We employ a progressive data reduction strategy: while keeping the test set unchanged, we systematically reduce the number of batteries in the training set for each specific operating condition to 20%, 40%, 60%, and 80% of the original training set (the complete training set contains 78 batteries, with the 20% subset corresponding to 21 batteries, following the rounding-up principle). This approach accurately simulates the data acquisition limitations commonly encountered in practical applications.

**Supplementary Note 11.** The proposed PIMOE framework for degradation trajectory prediction using historical capacity data.

In certain scenarios, battery recyclers may obtain partial historical maximum capacity data prior to battery retirement, which enables direct recycling without additional testing. Here, we further investigate the applicability and performance of the PIMOE model when directly predicting future degradation trajectories using existing maximum capacity data. Notably, since no within-cycle data is involved, we utilize historical capacity to drive the weight allocation in our degradation router, treating capacity magnitude and degradation trajectory as classification criteria for expert network assignment. Each expert network directly outputs degradation trends based on historical cycles rather than capacity-voltage curves. The historical data are decomposed via a feed-forward network (FFN) and fed into a noisy degradation router:

$$H(Q) = \mathbf{Q} \cdot \mathbf{W}_g + \psi \cdot \text{Softplus}(Q \cdot \mathbf{W}_{\text{noise}}) \tag{19}$$

Here, $\mathbf{Q}$ represents the historical cycle capacity data. The expert network weights $G(Q)$ are then derived from $\mathbf{Q}$ using the same methodology described in the Methods section. These adaptively integrated expert weights generate preliminary degradation trend predictions.

$$\boldsymbol{Trend_i} = \sum_{j=0}^{E} \mathbf{G}(\boldsymbol{Q_i}) \cdot \text{Predictor}_j(\mathbf{Q_i}) \tag{20}$$

The final degradation trajectory prediction is achieved by progressively incorporating future operating conditions using the identical FORNN architecture. Model prediction results are shown in Supplementary Fig. 24.

**Supplementary Table 1.** Technical Specifications of UL Dataset.

The table presents the material types, operating conditions, nominal capacities, cutoff voltages, and number of batteries used in our experimental dataset. All batteries in this study underwent identical load cycling throughout their full lifecycle until aging. The dataset is designated as the UL Dataset[4], and we adhere to the naming convention established by the data providers. For instance, the notation "NCA-45-05-1" refers to an NCA battery tested at 45°C with 0.5C CCCV charging and 1C discharging conditions, with similar naming logic applying to other cases. The UL Dataset encompasses 3 material types, 11 operating conditions, and 130 batteries in total. Notably, for each sample, we utilize only the current cycle's data without requiring any historical cycle information.

| Cell types | Working conditions | | | Nominal capacity (Ah) | Cut-off voltage (V) | Number of cells |
|---|---|---|---|---|---|---|
| | Charge C rate | Discharge C rate | Temperature (°C) | | | |
| NCA | 0.5 | 1 | 25 | 3.5 | 2.65-4.2 | 66 |
| | 0.5 | 1 | 35 | | | |
| | 0.5 | 1 | 45 | | | |
| | 1 | 1 | 25 | | | |
| | 0.25 | 1 | 25 | | | |
| NCM+NCA | 0.5 | 1 | 25 | 2.5 | 2.5–4.2 | 9 |
| | 0.5 | 2 | 25 | | | |
| | 0.5 | 4 | 25 | | | |
| NCM | 0.5 | 1 | 25 | 3.5 | 2.5-4.2 | 55 |
| | 0.5 | 1 | 35 | | | |
| | 0.5 | 1 | 45 | | | |

**Supplementary Table 2.** Technical Specifications of TPSL Dataset.

The table presents the materials, operating conditions, nominal capacities, cutoff voltages, and number of batteries used in our experimental dataset. After the initial 20 cycles, the load conditions were modified for these batteries, with 22 batteries undergoing different constant operating condition cycles and 55 batteries subjected to random operating condition cycles. The dataset is designated as the TPSL Dataset[24].

In this study, "TPSL-Arbitrary" refers to batteries operating under random conditions during their second-life usage, while "TPSL-Fixed" indicates batteries operating under fixed conditions during their second-life usage. The UL Dataset encompasses a total of 66 second-life operating conditions (55 random and 7 fixed) across 77 batteries. Notably, for each sample, we utilize only the current cycle's data without requiring any historical cycle information.

| Cell types | Working conditions | | | Nominal capacity (Ah) | Cut-off voltage (V) | Number of cells |
|---|---|---|---|---|---|---|
| | Charge C rate | Discharge C rate | Temperature (°C) | | | |
| NCM | 2 | 1 | 25 | 2.4 | 3.0-4.2 | 3 |
| | 3 | 1 | 25 | | | 3 |
| | 1 | 2 | 25 | | | 4 |
| | 2 | 2 | 25 | | | 3 |
| | 3 | 2 | 25 | | | 3 |
| | 2 | 3 | 25 | | | 3 |
| | 3 | 3 | 25 | | | 3 |
| | Random current (1C~3C) | 3 | 25 | | | 55 |

**Supplementary Table 3.** The model core configuration.

Here we show the hyperparameter configuration for our experiments. While more sophisticated training schemes and precise hyperparameter tuning might yield better results, we maintained identical hyperparameters across all datasets and operating conditions—except for the training length (set to 10 for NCA-25-1-1 dataset and 50 for others)—to ensure valid experimental verification through maximally simplified configurations.

| Level | Layer name | Parameter |
|---|---|---|
| AMDP | Expert Number | 5 |
| | Topk | 2 |
| | Degradation Router | $(I_S, O_S) = (12,5)$ |
| | $\alpha$ | 10 |
| | Predictor | $(I_S, O_S) = (50,50)$ |
| FORNN | hidden dim | 64 |
| | FFN | $(I_S, O_S) = (64,1)$ |
| | Dropout | 0.05 |
| | Activation function | ReLu |

**Supplementary Table 4.** The performance of the model in three different usage scenarios.

Here we demonstrates the predictive performance of the PIMOE model under two distinct scenarios: UL representing full lifecycle under constant operating conditions, and TPSL corresponding to repurposed usage with varying operating conditions (specific naming conventions are detailed in Supplementary Tabel 1). To ensure experimental rigor, all results represent averages from 10 randomized trials. Notably, the models exclusively utilized partial cycle data collected in situ as input to predict 50-cycle capacity degradation trajectories under uncertain future operating conditions.

| Error Metric | MAPE (%) | RMSE (%) | MAE (%) |
|---|---|---|---|
| UL | 0.520852 | 1.878831 | 1.404282 |
| TPSL-Arbitrary | 2.957535 | 4.277484 | 3.326660 |
| TPSL-Fixed | 2.806381 | 5.522538 | 3.857384 |

**Supplementary Table 5.** Model performance under 80% training data condition.

Here we addresses how limited data availability hinders the transition of data-driven approaches from laboratory to real-world deployment by evaluating PIMOE, PatchTST and Informer performance under constrained training data conditions. As a case study, we systematically reduce the training data to 80% of the original experimental dataset to examine the relationship between training data volume and model performance. All results represent average MAPE (%) from 10 randomized trials, with models exclusively using partial cycle data collected in situ to predict 50-cycle capacity degradation trajectories under uncertain future operating conditions.

| Model | PIMOE | PatchTST | Informer |
|---|---|---|---|
| NCACY45-05_1 | 0.38524 | 1.30522 | 1.32668 |
| NCACY35-05_1 | 0.54664 | 0.41313 | 0.65608 |
| NCACY25-025_1 | 0.99175 | 1.46143 | 1.20459 |
| NCACY25-05_1 | 1.1911 | 1.09639 | 1.13361 |
| NCACY25-1_1 | 1.27391 | 1.28571 | 1.30616 |
| NCMCY45-05_1 | 0.35657 | 0.36611 | 0.87559 |
| NCMCY35-05_1 | 0.45163 | 0.35729 | 0.63922 |
| NCMCY25-05_1 | 0.47977 | 0.30736 | 0.64768 |
| NCMNCACY25-05_1 | 0.3374 | 0.37809 | 0.50637 |
| NCMNCACY25-05_2 | 0.52637 | 0.52904 | 0.5344 |
| NCMNCACY25-05_4 | 0.35185 | 0.40742 | 0.57785 |
| TPSL-Arbitrary | 3.214 | 16.29637 | 16.3254 |
| TPSL-Fixed | 3.24584 | 30.6155 | 31.88632 |

**Supplementary Table 6.** Model performance under 60% training data condition.

Here we addresses how limited data availability hinders the transition of data-driven approaches from laboratory to real-world deployment by evaluating PIMOE, PatchTST and Informer performance under constrained training data conditions. As a case study, we systematically reduce the training data to 60% of the original experimental dataset to examine the relationship between training data volume and model performance. All results represent averaged MAPE (%) from 10 randomized trials, with models exclusively using partial cycle data collected in situ to predict 50-cycle capacity degradation trajectories under uncertain future operating conditions.

| Model | PIMOE | PatchTST | Informer |
| --- | --- | --- | --- |
| NCACY45-05_1 | 0.48513 | 1.37203 | 2.67657 |
| NCACY35-05_1 | 0.54664 | 0.41313 | 2.47614 |
| NCACY25-025_1 | 1.05483 | 1.70332 | 4.17579 |
| NCACY25-05_1 | 1.18969 | 1.12744 | 2.2514 |
| NCACY25-1_1 | 1.54336 | 1.35142 | 1.34329 |
| NCMCY45-05_1 | 0.38032 | 1.1135 | 1.16151 |
| NCMCY35-05_1 | 0.45163 | 0.35549 | 0.73698 |
| NCMCY25-05_1 | 0.49066 | 0.3395 | 1.46621 |
| NCMNCACY25-05_1 | 0.3374 | 0.37809 | 1.47611 |
| NCMNCACY25-05_2 | 0.52637 | 0.52904 | 1.47845 |
| NCMNCACY25-05_4 | 0.35185 | 0.40985 | 1.02613 |
| TPSL-Arbitrary | 3.20795 | 16.40015 | 16.16911 |
| TPSL-Fixed | 3.9755 | 30.95967 | 32.02418 |

**Supplementary Table 7.** Model performance under 40% training data condition.

Here we addresses how limited data availability hinders the transition of data-driven approaches from laboratory to real-world deployment by evaluating PIMOE, PatchTST and Informer performance under constrained training data conditions. As a case study, we systematically reduce the training data to 40% of the original experimental dataset to examine the relationship between training data volume and model performance. All results represent average MAPE (%) from 10 randomized trials, with models exclusively using partial cycle data collected in situ to predict 50-cycle capacity degradation trajectories under uncertain future operating conditions.

| Model | PIMOE | PatchTST | Informer |
| --- | --- | --- | --- |
| NCACY45-05_1 | 0.5128 | 1.42162 | 2.66275 |
| NCACY35-05_1 | 0.57415 | 0.57858 | 2.79688 |
| NCACY25-025_1 | 1.32399 | 1.65556 | 3.94724 |
| NCACY25-05_1 | 1.20611 | 1.11754 | 2.68264 |
| NCACY25-1_1 | 1.69018 | 1.50498 | 1.59012 |
| NCMCY45-05_1 | 0.44771 | 0.98612 | 1.36852 |
| NCMCY35-05_1 | 0.48234 | 0.43736 | 0.70738 |
| NCMCY25-05_1 | 0.55269 | 0.3646 | 1.82602 |
| NCMNCACY25-05_1 | 0.36781 | 0.53223 | 1.67119 |
| NCMNCACY25-05_2 | 0.68287 | 0.61551 | 1.64966 |
| NCMNCACY25-05_4 | 0.3993 | 0.54425 | 1.0907 |
| TPSL-Arbitrary | 3.39716 | 16.45436 | 16.31379 |
| TPSL-Fixed | 4.31718 | 31.22801 | 33.0205 |

**Supplementary Table 8.** Predicting the performance of different models over the next 30 cycles.

Here we examines the relationship between prediction horizon length and model performance in practical deployment scenarios. We systematically evaluate degradation trajectory predictions under uncertain future operating conditions with 30-cycle prediction lengths, while maintaining the model's input as partial current-cycle data. Notably, as previously mentioned, the NCACY25-1_1 dataset (with an average lifespan of approximately 30 cycles) retains a prediction length of 10 cycles. All results represent average MAPE (%) from 10 randomized trials to ensure experimental rigor.

| Model | PIMOE | PatchTST | Informer |
|---|---|---|---|
| NCACY45-05_1 | 0.34547 | 1.18713 | 1.35917 |
| NCACY35-05_1 | 0.41449 | 0.41208 | 0.59813 |
| NCACY25-025_1 | 1.16528 | 1.42827 | 1.06976 |
| NCACY25-05_1 | 0.92025 | 0.9871 | 1.09392 |
| NCACY25-1_1 | 0.97902 | 1.55313 | 1.96347 |
| NCMCY45-05_1 | 0.32087 | 0.23197 | 0.66038 |
| NCMCY35-05_1 | 0.6524 | 0.37717 | 0.64623 |
| NCMCY25-05_1 | 0.3936 | 0.30995 | 0.66665 |
| NCMNCACY25-05_1 | 0.38524 | 0.6007 | 0.58618 |
| NCMNCACY25-05_2 | 0.56414 | 0.6626 | 0.49081 |
| NCMNCACY25-05_4 | 0.42391 | 0.482 | 0.48673 |
| TPSL-Arbitrary | 2.18022 | 15.18396 | 15.20092 |
| TPSL-Fixed | 1.74537 | 19.34302 | 19.8816 |

**Supplementary Table 9.** Predicting the performance of different models over the next 80 cycles.

Here we examines the relationship between prediction horizon length and model performance in practical deployment scenarios. We systematically evaluate degradation trajectory predictions under uncertain future operating conditions with 80-cycle prediction lengths, while maintaining the model's input as partial current-cycle data. Notably, as previously mentioned, the NCACY25-1_1 dataset (with an average lifespan of approximately 30 cycles) retains a prediction length of 10 cycles. For the TPSL dataset (maximum 120 cycles), the 80-cycle prediction effectively constitutes an early-life prediction task. All results represent average MAPE (%) from 10 randomized trials to ensure experimental rigor.

| Model | PIMOE | PatchTST | Informer |
|---|---|---|---|
| NCACY45-05_1 | 0.5128 | 1.42162 | 2.66275 |
| NCACY35-05_1 | 0.57415 | 0.57858 | 2.79688 |
| NCACY25-025_1 | 1.32399 | 1.65556 | 3.94724 |
| NCACY25-05_1 | 1.20611 | 1.11754 | 2.68264 |
| NCACY25-1_1 | 1.69018 | 1.50498 | 1.59012 |
| NCMCY45-05_1 | 0.44771 | 0.98612 | 1.36852 |
| NCMCY35-05_1 | 0.48234 | 0.43736 | 0.70738 |
| NCMCY25-05_1 | 0.55269 | 0.3646 | 1.82602 |
| NCMNCACY25-05_1 | 0.36781 | 0.53223 | 1.67119 |
| NCMNCACY25-05_2 | 0.68287 | 0.61551 | 1.64966 |
| NCMNCACY25-05_4 | 0.3993 | 0.54425 | 1.0907 |
| TPSL-Arbitrary | 3.39716 | 16.45436 | 16.31379 |
| TPSL-Fixed | 4.31718 | 31.22801 | 33.0205 |

**Supplementary Table 10.** Predicting the performance of different models over the next 100 cycles.

Here we examines the relationship between prediction horizon length and model performance in practical deployment scenarios. We systematically evaluate degradation trajectory predictions under uncertain future operating conditions with 100 cycles prediction lengths, while maintaining the model's input as partial current-cycle data. Notably, as previously mentioned, the NCACY25-1_1 dataset (with an average lifespan of approximately 30 cycles) retains a prediction length of 10 cycles. For the TPSL dataset (maximum 120 cycles), the 100-cycle prediction effectively constitutes an early-life prediction task. All results represent average MAPE (%) from 10 randomized trials to ensure experimental rigor.

| Model | PIMOE | PatchTST | Informer |
| --- | --- | --- | --- |
| NCACY45-05_1 | 0.54324 | 0.95772 | 1.13189 |
| NCACY35-05_1 | 0.53951 | 0.51083 | 0.60636 |
| NCACY25-025_1 | 1.49899 | 1.51116 | 1.27905 |
| NCACY25-05_1 | 0.98466 | 1.00232 | 1.06487 |
| NCACY25-1_1 | 1.22211 | 1.55313 | 1.96347 |
| NCMCY45-05_1 | 0.39807 | 0.1943 | 0.72192 |
| NCMCY35-05_1 | 0.44048 | 0.36232 | 0.64616 |
| NCMCY25-05_1 | 0.4945 | 0.31077 | 0.55818 |
| NCMNCACY25-05_1 | 0.31185 | 0.3969 | 0.48392 |
| NCMNCACY25-05_2 | 0.60576 | 0.55978 | 0.49276 |
| NCMNCACY25-05_4 | 0.49869 | 0.49689 | 0.55327 |
| TPSL-Arbitrary | 5.31326 | 23.88182 | 23.88243 |
| TPSL-Fixed | 6.25851 | 72.2825 | 70.19593 |

**Supplementary Table 11.** Classification results based on expert-weighted evaluation for batteries at different SOH retirement scenarios.

| Condition | SOH | Total numbers | Battery Classification | | | Confidence Level |
|---|---|---|---|---|---|---|
| | | | Excellent | Qualified | Scrap | |
| NCACY45-05_1 | 95 | 28 | 27 | 1 | 0 | 96.4% |
| | 85 | 28 | 19 | 3 | 3 | / |
| | 75 | 28 | 1 | 1 | 26 | 92.9% |
| NCACY25-05_1 | 95 | 19 | 19 | 0 | 0 | 100% |
| | 85 | 19 | 5 | 5 | 9 | / |
| | 75 | 19 | 0 | 0 | 19 | 100% |
| NCACY25-1_1 | 95 | 9 | 9 | 0 | 0 | 100% |
| | 85 | 9 | 0 | 4 | 5 | / |
| | 75 | 9 | 0 | 0 | 9 | 100% |
| NCACY25-025_1 | 95 | 7 | 6 | 1 | 0 | 85.7% |
| | 85 | 7 | 0 | 5 | 2 | / |
| | 75 | 6 | 0 | 2 | 4 | 66.7% |
| NCACY35-05_1 | 95 | 3 | 3 | 0 | 0 | 100% |
| | 85 | 3 | 2 | 0 | 1 | / |
| | 75 | 2 | 0 | 0 | 2 | 100% |
| NCMCY45-05_1 | 95 | 28 | 19 | 9 | 0 | 67.9% |
| | 85 | 28 | 24 | 3 | 1 | / |
| | 75 | 23 | 0 | 0 | 23 | 100% |
| NCMCY35-05_1 | 95 | 4 | 4 | 0 | 0 | 100% |
| | 85 | 4 | 4 | 0 | 0 | / |
| | 75 | 4 | 0 | 0 | 4 | 100% |
| NCMCY25-05_1 | 95 | 23 | 23 | 0 | 0 | 100% |
| | 85 | 23 | 5 | 6 | 12 | / |
| | 75 | 7 | 0 | 0 | 7 | 100% |
| NCMNCACY25-05_1 | 95 | 3 | 3 | 0 | 0 | 100% |
| | 85 | 3 | 0 | 3 | 0 | / |
| | 75 | 3 | 0 | 0 | 3 | 100% |
| NCMNCACY25-05_2 | 95 | 3 | 3 | 0 | 0 | 100% |
| | 85 | 3 | 3 | 0 | 0 | / |
| | 75 | 3 | 0 | 0 | 3 | 100% |
| NCMNCACY25-05_4 | 95 | 3 | 3 | 0 | 0 | 100% |
| | 85 | 3 | 3 | 0 | 0 | / |
| | 75 | 3 | 0 | 0 | 3 | 100% |